\documentclass[sigconf]{acmart}

\copyrightyear{2025}
\acmYear{2025}
\setcopyright{acmlicensed}

\acmConference[CIKM '25] {Proceedings of the 34th ACM International Conference on Information and Knowledge Management}{ November 10--14, 2025}{Seoul, Republic of Korea.}

\acmBooktitle{Proceedings of the 34th ACM International Conference on Information and Knowledge Management (CIKM '25), November 10--14, 2025, Seoul, Republic of Korea}

\acmISBN{979-8-4007-2040-6/2025/11}

\acmDOI{10.1145/3746252.3761289}

\usepackage{xspace} 
\usepackage{threeparttable} 
\usepackage{enumitem} 
\usepackage{arydshln} 
\usepackage{subcaption} 

\usepackage{balance}

\newcommand{\paratitle}[1]{\vspace{1ex}\noindent{\bf #1}}

\newcommand{\ie}{\emph{i.e.,}\xspace}

\newcommand{\iid}{\emph{i.i.d.}\xspace}

\begin{document}

\newcommand{\cam}{CAM\xspace}
\newcommand{\fullcam}{Contextual Attention Modulation\xspace}

\newcommand{\hycam}{HyCAM\xspace}
\newcommand{\fullhycam}{Hybrid Contextual Attention Modulation\xspace}

\newcommand{\multilora}{Multi LoRA\xspace}
\newcommand{\moelora}{RieMoE-LoRA\xspace}

\title{Contextual Attention Modulation: Towards Efficient Multi-Task Adaptation in Large Language Models}

\author{Dayan Pan}
\affiliation{
  \institution{SCSE, Beihang University}
  \institution{ERC of ACAT, MOE}
  \city{Beijing}
  \country{China}
}
\affiliation{
  \institution{City University of Hong Kong}
  \city{Hong Kong}
  \country{China}
}
\email{dayan@buaa.edu.cn}

\author{Zhaoyang Fu}
\affiliation{
  \institution{Huawei Technologies Ltd.}
  \city{Shenzhen}
  \country{China}
}
\email{fuzhaoyang2@huawei.com}

\author{Jingyuan Wang}
\authornote{Corresponding authors.}
\affiliation{
  \institution{SCSE, Beihang University}
  \institution{Key Lab of DIM, MIIT}
  \institution{SEM, Beihang University}
  \city{Beijing}
  \country{China}
}
\email{jywang@buaa.edu.cn}

\author{Xiao Han}
\affiliation{
  \institution{Zhejiang University of Technology}
  \city{Hangzhou}
  \country{China}
}
\email{hahahenha@gmail.com}

\author{Yue Zhu}
\authornotemark[1]
\affiliation{
  \institution{Huawei Technologies Ltd.}
  \city{Shenzhen}
  \country{China}
}
\email{zhuyue9@huawei.com}

\author{Xiangyu Zhao}
\authornotemark[1]
\affiliation{
  \institution{City University of Hong Kong}
  \city{Hong Kong}
  \country{China}
}
\email{xianzhao@cityu.edu.hk}

\renewcommand{\shortauthors}{Dayan et al.}

\begin{abstract}
Large Language Models (LLMs) possess remarkable generalization capabilities but struggle with multi-task adaptation, particularly in balancing knowledge retention with task-specific specialization.
Conventional fine-tuning methods suffer from catastrophic forgetting and substantial resource consumption, while existing parameter-efficient methods perform suboptimally in complex multi-task scenarios.
To address this, we propose \fullcam (\cam), a novel mechanism that dynamically modulates the representations of self-attention modules in LLMs. \cam enhances task-specific features while preserving general knowledge, thereby facilitating more effective and efficient adaptation.
For effective multi-task adaptation, \cam is integrated into our \fullhycam (\hycam) framework, which combines a shared, full-parameter \cam module with multiple specialized, lightweight \cam modules, enhanced by a dynamic routing strategy for adaptive knowledge fusion.
Extensive experiments on heterogeneous tasks, including question answering, code generation, and logical reasoning, demonstrate that our approach significantly outperforms existing approaches, achieving an average performance improvement of 3.65\%. The implemented code and data are available to ease reproducibility.\footnote{https://github.com/Applied-Machine-Learning-Lab/HyCAM}

\end{abstract}

\ccsdesc[500]{Computing methodologies~Natural language processing}

\keywords{Large Language Model, Parameter-efficient fine-tuning
, Multi-Task Adaptation}


\maketitle
\section{Introduction} \label{sec:intro}
Large Language Models (LLMs) have demonstrated remarkable capabilities by their extensive general knowledge and powerful reasoning abilities~\cite{achiam2023gpt, team2023gemini}.
More than just a conversation, these models are increasingly proving invaluable as core components in advanced information retrieval~\cite{li2023e4srec, li2023web}, critical decision-making systems~\cite{brynjolfsson2025generative, wang2023rethinking}, and spatiotemporal applications~\cite{cheng2025poi, zhang2024veccity}.
The success has led to increasing demand for adapting such models to specialized domains and, more importantly, to handle multiple diverse tasks simultaneously.
This capability is essential for effective deployment in real-world applications~\cite{bommasani2021opportunities, yu2025bigcity, ji2025seeing}.

Supervised Fine-Tuning (SFT), a widely adopted adaptation approach, involves further tuning a pre-trained model on task-specific instruction data~\cite{wei2021finetuned}.
However, achieving effective adaptation remains significant challenges.
Conventional full parameter fine-tuning, a common SFT implementation that updates all parameters, needs to achieve effective adaptation while preserving foundational capabilities.
The training process on a narrow task-specific dataset can significantly change the model's pre-trained weights, leading to catastrophic forgetting~\cite{lester2021power}.
Furthermore, such an approach typically demands substantial computational resources.
Such limitations hinder its applicability in many practical scenarios, especially in multi-task settings~\cite{wang2023multi, fu2025training}, where models must balance between generalization and specialization.
To address these limitations, various Parameter-Efficient Fine-Tuning (PEFT) techniques have been proposed. These approaches adapt pre-trained LLMs to new tasks by updating only a small number of trainable parameters while leaving the backbone model unchanged, thereby reducing computational cost and overfitting risks~\cite{han2024parameter}.
Common PEFT strategies include adapter-based methods~\cite{houlsby2019parameter} that insert lightweight trainable modules, prompt-based methods such as Prefix Tuning~\cite{li2021prefix} that modify input representations, and reparameterization methods like Low-Rank Adaptation (LoRA)~\cite{hu2021lora} and its variants. LoRA, a widely utilized PEFT method, employs low-rank decomposition to weight updates, making it both efficient and effective.

However, these methods face limitations in complex multi-task scenarios due to their limited generalization and representational capacity across diverse tasks and potential interference when adapting to multiple objectives simultaneously~\cite{yu2020gradient, liu2021conflict, navon2022multi}.
Specifically for low-rank reparameterization approaches like LoRA, the low-rank adaptability may restrict model expressiveness when applied to highly complex tasks, resulting in suboptimal performance~\cite{pan2024lisa}.
While strategies like incorporating the Mixture-of-Experts (MoE) mechanism, which combines multiple specialized PEFT modules for multi-task adaptation, aim to enhance model capacity for diverse tasks, these MoE-based approaches can introduce additional challenges, including mitigating coupling effects and effectively managing the contributions of different experts~\cite{rajbhandari2022deepspeed}.

Overall, adapting LLMs to diverse tasks presents two major challenges: (1) preserving rich pretrained general knowledge while specializing for specific tasks, and (2) extending the multi-task capabilities of Parameter-Efficient Methods.

Our approach is motivated by a key observation regarding LLM architectures: different components in the Transformer reveal different roles and activation behaviors.
Existing literature suggests that Feed-Forward Network (FFN) layers, constituting the bulk of model parameters, primarily function as key repositories for storing and recalling general knowledge~\cite{geva2021transformer}.
In contrast, self-attention mechanisms are primarily responsible for processing and integrating contextual information within the input sequence, capturing dependencies between tokens~\cite{jin2025massive}.
This functional difference is also reflected in the parameter activation.
While FFNs, comprising approximately 90\% of model parameters, exhibit high activation sparsity, self-attention mechanisms typically demonstrate denser activation patterns~\cite{cai2024survey, fedus2022switch, jaszczur2021sparse}.
This denser engagement highlights its critical role in integrating latent general knowledge with contextual information derived from the input.

Given these differences, we argue that focusing on the modulation of self-attention during multi-task adaptation provides a more effective and specialized strategy.
The key insight is that large-scale pre-training has equipped LLMs with extensive general knowledge, so effective adaptation should focus on enabling LLMs to better integrate task-specific contextual information.
Such an approach can refine how general knowledge is integrated with specific contextual demands of diverse tasks. Importantly, this modulation preserves pre-trained general knowledge, thereby mitigating issues like catastrophic forgetting and task interference.

To this end, we introduce \fullcam (\cam), a novel mechanism designed to dynamically modulate the representations within the self-attention modules of LLMs based on the input context.
\cam learns to dynamically modulate self-attention representations to adapt the input context. 
This context-aware mechanism selectively amplifies task-relevant attentional signals and suppresses irrelevant or interfering ones, thereby enhancing task-specific features while preserving the model's pre-trained general knowledge.
Directly modulating the organization of contextual information within attention modules promotes more effective knowledge retention and specialized adaptation, thereby supporting more robust and efficient multi-task learning.

To extend the multi-task capabilities, we embed CAM into our Hybrid Contextual Attention Modulation (HyCAM) framework.
HyCAM combines a shared, full-parameter CAM module, which is designed to capture and leverage common knowledge across all tasks, with multiple specialized, lightweight CAM modules.
These specialized modules implement the CAM mechanism using PEFT techniques to efficiently capture distinct features, allowing effective multi-task adaptation with minimal additional trainable parameters.
A soft-routing strategy, further augmented by a load-balancing constraint, dynamically manages the fusion of knowledge from these shared and specialized CAM components.
This design empowers HyCAM to extend multi-task performance by enabling both efficient knowledge sharing and fine-grained specialization.

The main contributions of this paper are summarized as follows:
\begin{itemize}[leftmargin=*, topsep=0pt]
    \item We propose Contextual Attention Modulation (CAM), a novel mechanism that learns to dynamically modulate self-attention representations in LLMs based on input context. 
    CAM is designed to enhance task-specific features while preserving pre-trained general knowledge, thereby facilitating more effective knowledge retention and specialized adaptation.
    \item We introduce the Hybrid Contextual Attention Modulation (HyCAM) framework, which extends multi-task adaptation capabilities by integrating our CAM mechanism in distinct forms. This integration empowers HyCAM to achieve superior multi-task performance by effectively balancing efficient knowledge sharing with fine-grained task specialization.
    \item We conduct extensive experiments across a range of tasks covering question answering, code generation, logical reasoning, and other domains. Comparative experiments demonstrate that HyCAM significantly outperforms existing state-of-the-art approaches with faster convergence. 
\end{itemize}

\section{Preliminaries}
This section briefly reviews the fundamental concepts essential for understanding our proposed method. We discuss the relevant components of the Transformer architecture, the basics of task-adaptive fine-tuning, and common PEFT techniques.

\subsection{Transformer Architecture}
The Transformer architecture~\cite{vaswani2017attention} serves as the backbone of most LLMs owing to its ability to efficiently process sequences of data through attention mechanisms, making it especially powerful for understanding and generating human language.
A Transformer model is typically composed of a stack of identical blocks. 
Each block primarily contains two core components: the self-attention mechanism and the Feed-Forward Network (FFN). 
Self-Attention mechanism allows the model to weigh the importance of different tokens in an input sequence and capture contextual relationships by computing attention scores using Query ($Q$), Key ($K$), and Value ($V$) projections, often via scaled dot-product attention: $Attention(Q, K, V) = \text{softmax}\left(\frac{QK^T}{\sqrt{d_k}}\right)V$. Following this, the FFN, typically composed of two linear transformations with a non-linear activation, further processes each token's representation independently to express the complex knowledge of the model.

\subsection{Task-Adaptive Fine-Tuning} \label{sec:finetune}

While LLMs acquire extensive general knowledge and reasoning capabilities, they typically require further adaptation to specialize them for specific tasks and align their behavior with desired objectives, such as following instructions. 
A common approach for such task-adaptive fine-tuning is Supervised Fine-Tuning (SFT). In SFT, the model learns from examples that provide explicit input-output pairings. 
These pairings might illustrate a question with its corresponding answer or an instruction followed by the desired model output.
The primary goal is to adjust the model's parameters to minimize a task-specific loss function, such as cross-entropy loss for sequence generation or classification tasks.

\subsection{Parameter-Efficient Fine-Tuning}
Adapting LLMs to specific tasks often involves fine-tuning, but updating all parameters is computationally expensive. PEFT methods enable model adaptation by introducing a small set of new parameters or reparameterizing existing ones while keeping the backbone model weights frozen, significantly reducing computational costs.

A mainstream PEFT category is reparameterization, which introduces a smaller set of trainable parameters that efficiently influence the model's behavior.
For instance, a common strategy is to represent the change in a pre-trained weight matrix $W_0$ during adaptation as a low-rank update, based on the observation that task-specific changes often lie in a subspace of much lower dimensionality than the full parameter space.
Thus, instead of learning a large, dense update matrix $\Delta W$, these methods learn a low-rank approximation of it, such as $\Delta W = BA$, where $B$ and $A$ are much smaller matrices~\cite{hu2021lora}. 

\begin{figure*}[ht]
    \centering
    \includegraphics[width=0.75\linewidth]{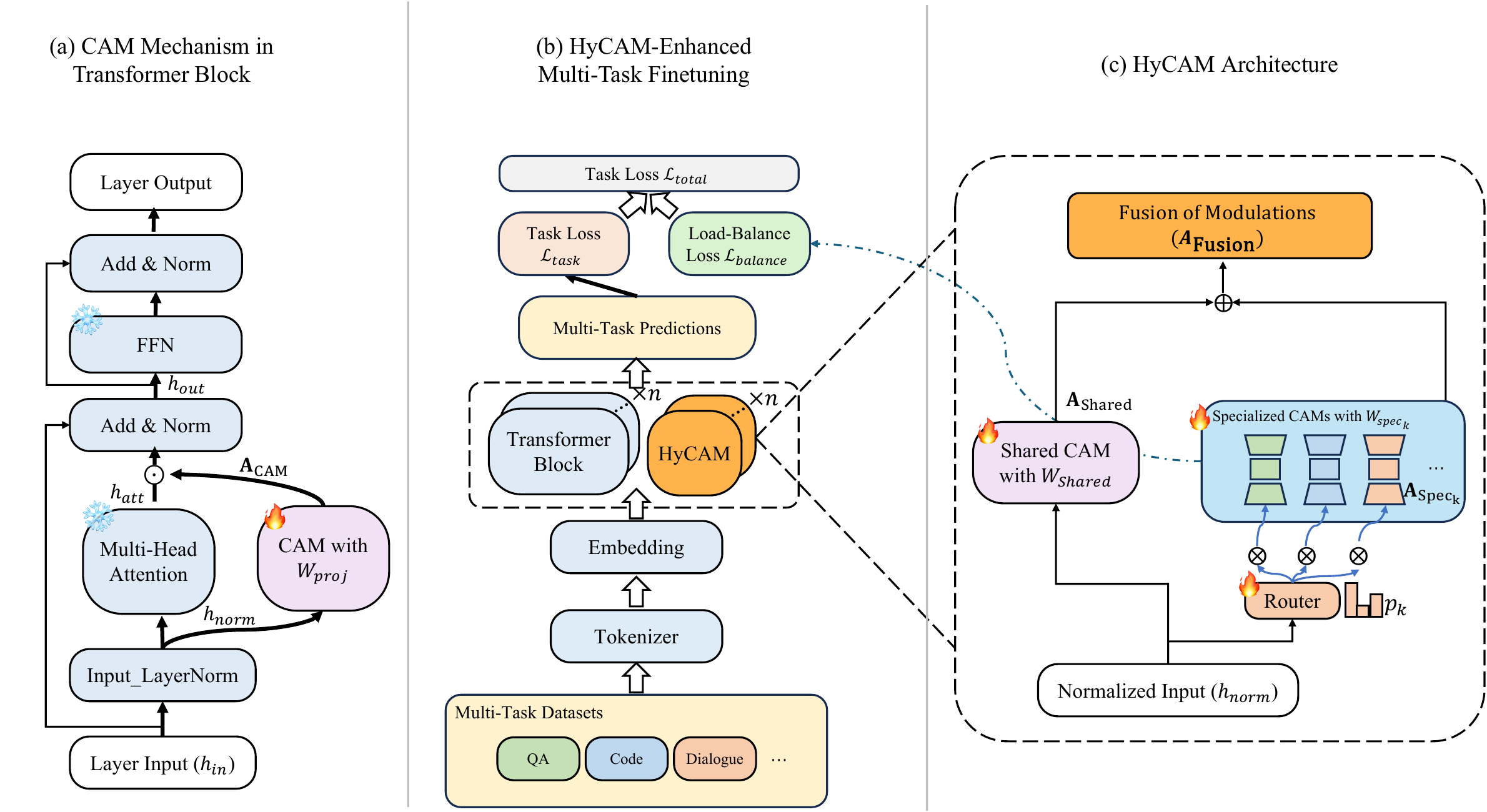}
    \caption{The architecture of the CAM and HyCAM framework. HyCAM applies a hybrid CAM mechanism to the output of the Attention module within each Transformer block, while the backbone LLM remains frozen. Specifically, HyCAM integrates a shared, full-parameter CAM module and multiple lightweight Specialized CAMs for common and task-specific knowledge.}
    \label{fig:model}
\end{figure*}

\section{Method}
We first illustrate an overview of our proposed HyCAM framework. 
Next, the core CAM mechanism is further detailed. We then provide an in-depth description of the HyCAM framework, including its hybrid components and dynamic knowledge fusion strategies with a soft-routing method and load-balancing constraint, and conclude by specifying the training objective.
\subsection{Framework Overview}
To address the critical challenge of enabling LLMs to efficiently adapt to diverse tasks while balancing knowledge retention with task-specific specialization, we introduce the Hybrid Contextual Attention Modulation (HyCAM) framework.
The core mechanism of HyCAM is Contextual Attention Modulation (CAM), which dynamically learns context-dependent modulation of self-attention representations, selectively amplifying task-relevant signals while suppressing irrelevant or potentially interfering ones to enhance task-specific features and preserve general knowledge.
As illustrated in Figure~\ref{fig:model}, the HyCAM framework employs a novel hybrid architecture that integrates a shared, full-parameter CAM module, designed for capturing common knowledge across tasks, with multiple specialized CAM modules that utilize parameter-efficient techniques for efficient, fine-grained adaptation to distinct task features.
The contributions of these diverse CAM modules are managed by a dynamic routing strategy to ensure balanced utilization of the specialized components and adaptive knowledge fusion.

\subsection{Contextual Attention Modulation} \label{sec:cam}
The CAM mechanism is the core of our HyCAM framework, designed to dynamically modulate self-attention representations at each Transformer block.
It learns to dynamically amplify task-relevant attentional signals and suppress irrelevant ones based on the input context, thereby enhancing task-specific features while preserving the model's pre-trained general knowledge, which facilitates more effective and efficient task adaptation.

\subsubsection{\textbf{Motivation}}
Our motivation for developing CAM comes from the analysis of distinct roles and activation patterns across different Transformer components, as described in Section~\ref{sec:intro}.
While FFN modules account for a large portion of parameters and store a vast amount of an LLM's parameterized knowledge, self-attention modules are crucial for dynamically processing and integrating contextual information.
The varying activation patterns of these components highlight the important role of the self-attention modules in integrating latent general knowledge with the specific context derived from an input.
With extensive general knowledge from large-scale pre-training of LLMs, the key to effective adaptation lies in enabling them to better integrate this foundational knowledge with task-specific contextual information.
Conventional fine-tuning approaches, however, can often overwrite the valuable pre-trained representations by introducing new task-specific knowledge.

This observation motivated us to develop CAM, a mechanism that refines how general knowledge is integrated with specific contextual demands of diverse tasks by modulating self-attention representations. This approach aims to facilitate task-adaptive specialization while preserving valuable pre-trained knowledge.

\subsubsection{\textbf{The CAM Mechanism}} \label{subsec:camdetail}
The CAM mechanism is integrated into each Transformer block, operating on the output of the self-attention modules to dynamically modulate its representations based on the input context. This process allows for a fine-grained modulation of contextual information flow.
Specifically, the CAM mechanism proceeds as follows:

\paratitle{Input Normalization: }
Let $h_{in} \in \mathbb{R}^{L \times d}$ be the input hidden state to a Transformer layer, where $L$ denotes the sequence length and $d$ represents the hidden dimension.
Consistent with standard Transformer operations, these input hidden states are first normalized using Layer Normalization~\cite{ba2016layer}, producing $h_{norm} \in \mathbb{R}^{L \times d}$:
\begin{equation}
    h_{norm} = \text{LayerNorm}(h_{in}).
\end{equation}  
The resulting $h_{norm}$ serves as the input for both the conventional self-attention computation and our CAM module.

\paratitle{Modulation Weight Generation: }
CAM then computes a context-dependent modulation weight tensor, denoted as $\mathbf{A}_{\text{CAM}} \in \mathbb{R}^{L \times d}$.
These weights are derived from the normalized hidden state $h_{norm}$ through a linear projection parameterized by a trainable weight matrix $W_{proj} \in \mathbb{R}^{d \times d}$, followed by a SiLU activation function~\cite{elfwing2018sigmoid}:
\begin{equation}
    \mathbf{A}_{\text{CAM}} = \text{SiLU}(h_{norm} W_{proj}).
\end{equation}    
    The matrix $W_{proj}$ is specific to the CAM module and is crucial for learning how to modulate the attention representations based on the input context.
    To ensure stability during the initial phases of fine-tuning and to allow the model to gradually learn the modulation, $W_{proj}$ is initialized as a zero matrix.
    This initialization strategy ensures that at the beginning of the fine-tuning, CAM does not alter the pre-trained model's behavior. That is, the model initially maintains its original approach to processing contextual information, which is then gradually modulated as training progresses for a stable adaptation.

\paratitle{Application of Modulation: }
Concurrently, the standard attention output $h_{att} \in \mathbb{R}^{L \times d}$ is computed using the normalized input $h_{norm}$:
\begin{equation}
    h_{att} = \text{Self-Attention}(h_{norm}).
    \label{eq:oriattn}
\end{equation}    
    The CAM mechanism then refines this $h_{att}$ by applying the learned modulation weights $\mathbf{A}_{\text{CAM}}$. This is performed via an element-wise Hadamard product ($\odot$). 
    The modulated signal is integrated with the original $h_{att}$ through a residual connection, forming the final output $h_{out} \in \mathbb{R}^{L \times d}$ of the attention mechanism incorporating CAM.
\begin{equation}
    h_{out} = h_{att} + h_{att} \odot \mathbf{A}_{\text{CAM}}.
\end{equation}

\subsubsection{\textbf{Advantages}}
    By dynamically generating and applying these modulation weights, CAM refines the contextual representation from the self-attention modules to adapt it to specific tasks while preserving the pre-trained general knowledge, thereby mitigating catastrophic forgetting.
    Thus, CAM facilitates an effective balance between achieving task-specific adaptation and retaining extensive general knowledge.
    Moreover, by modulating attentional outputs instead of fine-tuning a large number of backbone parameters, CAM achieves computational efficiency.

\vspace{-5px}
\subsection{The HyCAM Framework}
While the CAM mechanism provides a powerful tool for modulating attention representations,
adapting LLMs to handle multiple diverse tasks simultaneously presents significant challenges.
Conventional full fine-tuning struggles with catastrophic forgetting and resource demands, while existing PEFT methods still face limitations for multi-tasking. 
Specifically, the limited capacity of representation makes it suboptimal for highly complex tasks, and simple applications of expert-based strategies lead to an imbalance in expert utilization.

To address these multiple challenges and effectively leverage the CAM mechanism for complex multi-task learning scenarios, we introduce the HyCAM framework. 
The framework is designed to extend the multi-task adaptation capabilities by integrating CAM in hybrid forms, enabling both efficient knowledge sharing and fine-grained task specialization.
This is achieved through strategically combining shared, full-parameter CAM module, for efficient knowledge sharing, with multiple specialized, parameter-efficient CAM modules, for fine-grained specialization.
The contributions of these components are coordinated by a dynamic routing mechanism with a load-balancing constraint to ensure adaptive knowledge fusion.

\subsubsection{\textbf{Hybrid CAM Components}}
The hybrid architecture of the HyCAM framework is designed to leverage both general context understanding and specialized, task-specific adaptation capabilities. This architecture comprises a shared, full-parameter CAM module and multiple lightweight, specialized CAM modules: 

 \paratitle{Shared CAM Module: }
  The Shared CAM module serves as a global modulator, for capturing and refining common contextual patterns and general knowledge across all tasks. This module is a full-parameter CAM, as detailed in Section~\ref{sec:cam}. Its trainable projection matrix, denoted as $W_{Shared} \in \mathbb{R}^{d \times d}$, is shared and updated across all tasks to produce a modulation weight tensor:
\begin{equation} 
  \mathbf{A}_{Shared} = \text{SiLU}(h_{norm}W_{Shared}).
\end{equation}

\paratitle{Specialized CAM Modules: }
In addition to the shared module, HyCAM incorporates multiple ($N_s$) lightweight Specialized CAM modules.
Specialized CAM modules are designed to learn and apply attention modulations for the distinct features of specific tasks.

Different tasks often require different ways of handling contextual information in the self-attention layer. For example, code generation may need to focus on long-range dependencies, while question answering systems may prioritize specific entities and their relationships in a local context. This design is to enable the model to develop fine-grained adaptations for 
diverse tasks, thereby mitigating the interference when a single component attempts to learn potentially conflicting objectives from multiple tasks.

The implementation of Specialized CAM modules leverages the PEFT technique for reducing the number of trainable parameters per specialized module, making the framework scalable.
Besides, it helps in mitigating overfitting, especially when task-specific data might be limited.
Specifically, each Specialized CAM module, indexed by $k \in \{1, ..., N_s\}$, generates its unique modulation weight tensor $\mathbf{A}_{\text{Spec}_k} \in \mathbb{R}^{L \times d}$ as follows:
\begin{equation}
    \mathbf{A}_{\text{Spec}_k} = \text{SiLU}(h_{norm} W_{\text{Spec}_k}),
\end{equation}
where $W_{\text{Spec}_k}$ is the trainable projection matrix specific to the $k$-th specialized module. To achieve parameter efficiency while enhancing representational capacity, we adopt the SLoRA~\cite{guo2025nlora} technique for the structure of $W_{Spec_k}$. Instead of a direct low-rank decomposition like LoRA, typically $W = BA$, SLoRA introduces an intermediate trainable matrix $N$ between $B$ and $A$. Thus, $W_{Spec_k}$ is parameterized as:
\begin{equation}
    W_{Spec_k} = B_k N_k A_k.
\end{equation}
Here, $A_k \in \mathbb{R}^{r \times d}$ is a matrix that projects the $d$-dimensional hidden state $h_{norm}$ into a lower-dimensional space of rank $r$. $N_k \in \mathbb{R}^{r \times r}$ is a trainable intermediate matrix within the low-rank space. 
$B_k \in \mathbb{R}^{d \times r}$ is a matrix that projects the $r$-dimensional representation back to the original $d$-dimensional space.
The rank $r$ is significantly smaller than $d$, ensuring a substantial reduction in trainable parameters compared to a full $d \times d$ matrix.

For initialization, and similar to the zero-initialization of $W_{\text{Shared}}$ in the Shared CAM module, we adopt a strategy to ensure training stability. Specifically, the matrices $A_k$ and $N_k$ are initialized using Kaiming Uniform~\cite{he2015delving}. The matrix $B_k$ is initialized with zeros. This structure allows each Specialized CAM to develop task-specific modulations with very small parameters, thus enhancing the adaptability of the model without sacrificing efficiency.

\subsubsection{\textbf{Dynamic Routing}} \label{sec:routing}
To effectively leverage the diverse contributions from the Shared CAM and multiple Specialized CAM modules, HyCAM incorporates a dynamic soft-routing mechanism coupled with a load-balancing constraint.
This mechanism adaptively determines the influence of each module based on the input context and promotes load-balance to ensure efficient utilization of all Specialized CAMs.

\paratitle{Routing for Specialized CAMs: }
The dynamic routing mechanism weights the contributions of the $N_s$ Specialized CAM modules for each input token. This enables HyCAM to adapt its modulation strategy in a fine-grained, context-dependent manner. The routing process is detailed as follows:

For each token representation $h_{norm}$, derived from $h_{in}$ as described in Section~\ref{subsec:camdetail}, a lightweight router network first generates $\mathbf{logits} \in \mathbb{R}^{N_s}$, produced by a linear layer applied to $h_{norm}$:
\begin{equation}
    \mathbf{logits} = h_{norm} W_{router},
\end{equation}
where $W_{router} \in \mathbb{R}^{d \times N_s}$ is the trainable weight matrix of the router.

These  $\mathbf{logits}= [\pi_1, \pi_2, ..., \pi_{N_s}]$ are then transformed into a probability distribution over the specialized modules using the Gumbel-Softmax estimator~\cite{jang2016categorical} to obtain differentiable, soft routing probabilities.
The Gumbel-Softmax allows for differentiable sampling from a categorical distribution, which facilitates the training process while encouraging exploration, as detailed:
\begin{equation}
    p_k = \frac{\exp((\log \pi_k + g_k)/\tau)}{\sum_{j=1}^{N_s} \exp((\log \pi_j + g_j)/\tau)},
    \label{eq:gumbel_softmax}
\end{equation}
where $p_k$ is the resulting soft routing weight for the $k$-th Specialized CAM module. $g_k \sim \text{Gumbel}(0,1)$ are \iid noise drawn from the Gumbel distribution, adding stochasticity for exploration. $\tau$ is a temperature hyperparameter that controls the sharpness of the probability distribution. Lower temperatures make the selection more discrete, while higher temperatures make it softer. 

\paratitle{Load Balancing Loss: }
To prevent routers from over-selecting a few modules, HyCAM adds a load-balancing loss $\mathcal{L}_{balance}$ that encourages more balanced routing across specialized components. For a batch of $B$ tokens, it is computed as:

\begin{equation}
    \mathcal{L}_{balance} =  \sum_{k=1}^{N_s} \left( \frac{1}{B} \sum_{b=1}^{B} p_{b,k} \right) \cdot \left( \frac{1}{B} \sum_{b=1}^{B} \text{softmax}(\mathbf{logits}_{b})_k \right),
    \label{eq:load_balance_loss}
\end{equation}
where $p_{b,k}$ is the Gumbel-Softmax output and $\text{softmax}(\mathbf{logits}_{b})_k$ is the standard softmax output of the router logits. 

\paratitle{Fusion of Modulations: }
Once the routing weights $p_k$ are determined for each token, as described in Equation~\ref{eq:gumbel_softmax}, the final context-dependent modulation tensor, $\mathbf{A}_{Fusion} \in \mathbb{R}^{L \times d}$, is computed by combining the output of the Shared CAM module, $\mathbf{A}_{Shared}$, with the dynamically weighted sum of the modulations from all Specialized CAM modules, $\{\mathbf{A}_{Spec_k}\}_{k=1}^{N_s}$:
\begin{equation}
    \mathbf{A}_{Fusion} = \mathbf{A}_{Shared} + \sum_{k=1}^{N_s} p_k \cdot \mathbf{A}_{Spec_k},
    \label{eq:fusion_modulation}
\end{equation}
Here, $p_k$ denotes the token-specific routing weight of the $k$-th specialized module, ensuring that the context-based modulation of $\mathbf{A}_{Fusion}$ integrates both general and adaptively selected specialized knowledge.
Finally, it is applied to the original self-attention output $h_{att}$, from Equation~\ref{eq:oriattn} in Section~\ref{subsec:camdetail}, to produce the HyCAM-enhanced output $h_{out}$ using the element-wise Hadamard product and residual connection, as defined in the core CAM mechanism:
\begin{equation}
    h_{out} = h_{att} + h_{att} \odot \mathbf{A}_{Fusion}.
\end{equation}
This entire mechanism, from dynamic routing to the application of the fused modulation,  allows HyCAM to dynamically modulate the self-attention process by integrating shared knowledge with specialized insights, thereby enabling the model to effectively balance generalization across diverse tasks with task-specific adaptation.

\begin{table*}[t] 
\small
    \centering
    \caption{Datasets statistics.}
    \label{tab:dataset}
    \resizebox{0.85\linewidth}{!}{
    \renewcommand{\arraystretch}{0.94}
    \begin{threeparttable}[b]
    \begin{tabular}{lccccc}
        \toprule
        Dataset     & Samples     & Total Tokens\tnote{1} & Avg. Tokens/Sample\tnote{1}  & Domain & Source \\
        \midrule
        Auto CoT        & 5,816          &   943,474                 & 162.22 & Arithmetic and other logical reasoning tasks & \cite{zhang2023automatic}\\
        iCliniq        & 7,321          &   1,826,306                & 249.46 & Conversations between patients and doctors & \cite{li2023chatdoctor}\\
        Dolly 2.0       & 15,015         &   3,061,007                 & 203.86 & Closed QA and summarization from Wikipedia & \cite{DatabricksBlog2023DollyV2}\\
        CodeAlpaca      & 20,222         &   2,195,523                 & 109.66 & Code generation and optimization & \cite{codealpaca}\\
        WebGPT          & 18,994         &   13,988,895                & 736.49 & Information retrieval QA & \cite{nakano2021webgpt}\\
        \bottomrule
    \end{tabular}
    \begin{tablenotes}
        \item[1] Calculated by Llama-3 Tokenizer.
    \end{tablenotes}
    \end{threeparttable}
    }
\end{table*}

\begin{table*}[t]
    \centering
    \caption{Experimental results across different backbone LLMs.
    \textbf{*}indicates the statistically significant improvements (\ie two-sided t-test with $p<0.05$) over the best PEFT baseline. Lower PPL$\downarrow$  is better, where higher BLEU$\uparrow$ and ROUGE$\uparrow$ reflect higher quality. The best results are bolded, while the second-best results are underlined.
    }
    \label{tab:exp1}
    \resizebox{0.95\linewidth}{!}{
    \renewcommand{\arraystretch}{1}
    \begin{tabular}{l|ccc|ccc|ccc|ccc|ccc}
        \toprule
        Backbone LLM & \multicolumn{3}{c|}{Llama 2 7B}& \multicolumn{3}{c|}{Llama 3 8B} & \multicolumn{3}{c|}{Llama 3.1 8B}   & \multicolumn{3}{c|}{Mistral 7B} & \multicolumn{3}{c}{Qwen 2.5 7B} \\ \midrule
        Metric  & PPL$\downarrow$   & BLEU$\uparrow$    & ROUGE$\uparrow$   & PPL$\downarrow$   & BLEU$\uparrow$ & ROUGE$\uparrow$ & PPL$\downarrow$  & BLEU$\uparrow$ & ROUGE$\uparrow$& PPL$\downarrow$  & BLEU$\uparrow$ & ROUGE$\uparrow$& PPL$\downarrow$  & BLEU$\uparrow$ & ROUGE$\uparrow$\\ \midrule
Full Finetune   & 3.193 & \underline{0.171} & 0.231 & 3.978 & 0.151 & 0.203 & 3.873 & 0.153 & 0.205 & 4.403 & 0.157 & 0.192 & 3.024 & \underline{0.169} & 0.225 \\
LoRA            & 3.222 & 0.157 & 0.225 & 3.556 & 0.148 & 0.24 & 3.537 & 0.156 & 0.237 & \underline{3.418} & \underline{0.163} & \underline{0.244} & 2.840 & 0.137 & \underline{0.239} \\
\midrule
\multilora      & 3.287 & 0.121 & 0.217 & 3.547 & 0.157 & 0.236 & 3.653 & 0.134 & 0.235 & 3.461 & 0.141 & 0.225 & 3.069 & 0.136 & 0.222 \\
\moelora        & \underline{3.171} & 0.154 & \underline{0.232} & \underline{3.497} & \underline{0.159} & \underline{0.242} & \underline{3.487} & \underline{0.161} & \underline{0.238} & 3.597 & 0.143 & 0.24 & \underline{2.830} & 0.157 & 0.227 \\
HyCAM           & \textbf{3.081*} & \textbf{0.173*} & \textbf{0.244*} & \textbf{3.484*} & \textbf{0.162*} & \textbf{0.245*} & \textbf{3.453*} & \textbf{0.172*} & \textbf{0.251*} & \textbf{3.299*} & \textbf{0.171*} & \textbf{0.249*} & \textbf{2.757*} & \textbf{0.172*} & \textbf{0.248*} \\
        \bottomrule
    \end{tabular}
    }
\end{table*}

\subsection{Training Details}
The HyCAM framework, including the Shared CAM module, the Specialized CAM modules, and the dynamic router, is trained end-to-end.
We use a composite objective function that combines a primary task-specific loss with the auxiliary load-balancing loss, described in Section~\ref{sec:routing}.
This approach ensures that the model not only learns to perform the target tasks effectively but also maintains balanced utilization of its specialized components, leading to efficient adaptation across diverse tasks and enhanced overall multi-task performance.

\paratitle{Task-specific Loss: }
We employ a standard autoregressive training strategy common for LLMs, as introduced in Section~\ref{sec:finetune}, where the model is trained to predict the next token in a sequence given the input context.
Given an input sequence $\mathbf{X} = (x_1, x_2, \dots, x_m)$ and its corresponding target sequence $\mathbf{Y} = (y_1, y_2, \dots, y_n)$, the model is trained to predict each token $y_t$ conditioned on the input $\mathbf{X}$ and the previous target tokens $\mathbf{Y}_{<t} = (y_1, y_2, \dots, y_{t-1})$. The primary objective is to minimize the cross-entropy loss of the target sequences:
\begin{equation}
    \mathcal{L}_{\text{task}} = -\sum_{i=1}^{|\mathcal{D}|} \sum_{t=1}^{n_i} \log P(y_{i,t} | \mathbf{X}_i, \mathbf{Y}_{i,<t}; \Theta_\text{HyCAM}),
    \label{eq:sft_loss}
\end{equation}
where $\mathcal{D}$ represents the batch of training examples, $n_i$ is the length of the $i$-th target sequence, and $P(y_{i,t} | \mathbf{X}_i, \mathbf{Y}_{i,<t}; \Theta_{HyCAM})$ is the probability of the true token $y_{i,t}$ predicted by the HyCAM framework with its trainable parameters $\Theta_\text{HyCAM}$. 

\paratitle{Overall Training Objective: }
To ensure diverse utilization of the Specialized CAM modules, we incorporate the auxiliary load-balancing loss $\mathcal{L}_{\text{balance}}$, as defined in Equation~\ref{eq:load_balance_loss}, into the overall training objective. The final training loss $\mathcal{L}_{\text{total}}$ that HyCAM optimizes is therefore a weighted sum of the task loss and the load-balancing loss:
\begin{equation}
    \mathcal{L}_{\text{total}} = \mathcal{L}_{\text{task}} + \lambda_{\text{balance}} \cdot \mathcal{L}_{\text{balance}}.
    \label{eq:total_loss}
\end{equation}
Here, $\lambda_{\text{balance}}$ is a hyperparameter that controls the contribution of the load-balancing constraint.
By optimizing $\mathcal{L}_{\text{total}}$, the HyCAM framework learns to effectively perform the target tasks while maintaining a balanced and efficient use of its specialized components.

\begin{table*}[t]
    \centering
    \caption{Results with different sizes of Qwen 2.5-Family.}
    \label{tab:expdiffqwen}
    \resizebox{0.95\linewidth}{!}{
    \renewcommand{\arraystretch}{1}
    \begin{tabular}{l|ccc|ccc|ccc|ccc|ccc}
        \toprule
        Backbone & \multicolumn{3}{c|}{Qwen 2.5 0.5B}& \multicolumn{3}{c|}{Qwen 2.5 1.5B} & \multicolumn{3}{c|}{Qwen 2.5 3B}   & \multicolumn{3}{c|}{Qwen 2.5 7B} & \multicolumn{3}{c}{Qwen 2.5 14B} \\ \midrule
        Metric & PPL$\downarrow$  & BLEU$\uparrow$ & ROUGE$\uparrow$ & PPL$\downarrow$  & BLEU$\uparrow$ & ROUGE$\uparrow$ & PPL$\downarrow$  & BLEU$\uparrow$ & ROUGE$\uparrow$& PPL$\downarrow$  & BLEU$\uparrow$ & ROUGE$\uparrow$& PPL$\downarrow$  & BLEU$\uparrow$ & ROUGE$\uparrow$\\ \midrule
Full Finetune & 3.778 & \underline{0.159} & 0.219 & \textbf{3.102} & \textbf{0.169} & \underline{0.235} & \underline{2.982} & \underline{0.161} & 0.222 & 3.024 & \underline{0.169} & 0.225 & 2.839 & \textbf{0.176} & 0.214 \\
LoRA & 3.764 & 0.145 & 0.222 & 3.344 & 0.138 & 0.229 & 3.106 & 0.144 & 0.230 & 2.840 & 0.137 & \underline{0.239} & 2.889 & 0.147 & \underline{0.238} \\
\midrule
\multilora & 3.754 & 0.144 & 0.221 & 3.330 & 0.148 & 0.226 & 3.053 & 0.157 & 0.225 & 3.069 & 0.136 & 0.222 & 2.882 & 0.152 & 0.235 \\
\moelora & \underline{3.621} & 0.152 & \underline{0.232} & 3.180 & 0.148 & 0.230 & 3.001 & 0.148 & \underline{0.238} & \underline{2.83} & 0.157 & 0.227 & \underline{2.792} & 0.142 & \underline{0.238} \\
HyCAM & \textbf{3.611} & \textbf{0.169} & \textbf{0.262} & \underline{3.108} & \underline{0.167} & \textbf{0.236} & \textbf{2.940} & \textbf{0.165} & \textbf{0.249} & \textbf{2.757} & \textbf{0.172} & \textbf{0.248} & \textbf{2.682} & \underline{0.160} & \textbf{0.242} \\
        \bottomrule
    \end{tabular}
    }
\end{table*}

\begin{table}[t]
    \centering
    \caption{Results with different sizes of Llama 3.2.}
    \label{tab:expsmallllama}
    \resizebox{0.95\linewidth}{!}{
    \renewcommand{\arraystretch}{1}
    \begin{tabular}{l|ccc|ccc|ccc|ccc|ccc}
        \toprule
        Backbone & \multicolumn{3}{c|}{Llama 3.2 1B}& \multicolumn{3}{c|}{Llama 3.2 3B}\\ \midrule
        Metric & PPL$\downarrow$  & BLEU$\uparrow$ & ROUGE$\uparrow$& PPL$\downarrow$  & BLEU$\uparrow$ & ROUGE$\uparrow$\\ \midrule
Full Finetune & \textbf{4.221} & \textbf{0.164} & 0.221 & \textbf{3.747} & \underline{0.159} & 0.220 \\
LoRA & 4.515 & 0.144 & 0.227 & 3.824 & 0.144 & \underline{0.234} \\
\midrule
\multilora & 4.533 & 0.143 & 0.225 & 3.876 & 0.149 & 0.232 \\
\moelora & 4.324 & 0.161 & \underline{0.241} & 3.806 & 0.154 & 0.233 \\
HyCAM & \underline{4.227} & \underline{0.163} & \textbf{0.244} & \underline{3.778} & \textbf{0.167} & \textbf{0.243} \\
        \bottomrule
    \end{tabular}
    }
\end{table}

\begin{table*}[t]
    \centering
    \caption{Detailed experimental results across the five datasets.}
    \label{tab:crosstaskresult}
    \resizebox{0.95\linewidth}{!}{
    \renewcommand{\arraystretch}{1}
    \begin{tabular}{l|ccc|ccc|ccc|ccc|ccc}
        \toprule
        Backbone LLM & \multicolumn{3}{c|}{
        Auto CoT}& \multicolumn{3}{c|}{iCliniq} & \multicolumn{3}{c|}{Dolly 2.0}   & \multicolumn{3}{c|}{CodeAlpaca} & \multicolumn{3}{c}{WebGPT} \\ \midrule
        Metric & PPL$\downarrow$  & BLEU$\uparrow$ & ROUGE$\uparrow$ & PPL$\downarrow$  & BLEU$\uparrow$ & ROUGE$\uparrow$ & PPL$\downarrow$  & BLEU$\uparrow$ & ROUGE$\uparrow$& PPL$\downarrow$  & BLEU$\uparrow$ & ROUGE$\uparrow$& PPL$\downarrow$  & BLEU$\uparrow$ & ROUGE$\uparrow$\\ \midrule
Full Finetune & 1.842 & \underline{0.282} & 0.287 & \textbf{7.497} & \textbf{0.053} & 0.123 & 6.461 & 0.088 & \textbf{0.200} & 2.532 & 0.138 & 0.195 & \underline{1.888} & \textbf{0.182} & \textbf{0.341} \\
LoRA & 1.843 & 0.268 & 0.291 & 8.140 & 0.049 & \underline{0.124} & 6.029 & 0.070 & 0.181 & 2.404 & \underline{0.146} & 0.202 & 1.919 & 0.178 & 0.331  \\
\midrule
\multilora & 1.952 & 0.198 & 0.290 & 8.846 & 0.037 & 0.122 & \textbf{5.743} & 0.101 & 0.177 & \textbf{2.312} & 0.134 & 0.189 & 1.939 & 0.176 & \underline{0.337} \\
\moelora & \underline{1.813} & 0.275 & \textbf{0.298} & 8.001 & 0.051 & 0.123 & 5.954 & \textbf{0.106} & 0.183 & 2.381 & 0.142 & \underline{0.207} & \underline{1.888} & 0.177 & 0.336  \\
HyCAM & \textbf{1.777} & \textbf{0.283} & \underline{0.297} & \underline{7.546} & \textbf{0.053} & \textbf{0.125} & \underline{5.893} & \underline{0.093} & \underline{0.194} & \underline{2.359} & \textbf{0.163} & \textbf{0.222} & \textbf{1.845} & \underline{0.180} & \underline{0.337} \\

        \bottomrule
    \end{tabular}
    }
\end{table*}

\section{Experiments}
In this section, we present a comprehensive evaluation of our proposed Hybrid Contextual Attention Modulation (HyCAM) framework.
To systematically evaluate the performance and characteristics of HyCAM, we conduct experimental analysis around the following key research questions (RQs):
\begin{itemize}[leftmargin=*]
    \item \textbf{RQ1}: How does the HyCAM framework perform overall in multi-task adaptation compared to state-of-the-art baseline methods across various backbone models?
    \item \textbf{RQ2}: How does HyCAM scale when applied to backbone Large Language Models of different sizes?
    \item \textbf{RQ3}: How does HyCAM perform on individual tasks within the multi-task benchmark, and does it demonstrate a balanced adaptation capability across these diverse task types?
    \item \textbf{RQ4}: What are the contributions of the core CAM mechanism and other components of HyCAM to its overall effectiveness, and how sensitive is its performance to key hyperparameters?
    \item \textbf{RQ5}: What qualitative insights do visualizations offer regarding HyCAM's internal working mechanisms?
\end{itemize}

\subsection{Experimental Setup}
\subsubsection{\textbf{Datasets}}
To better evaluate the effectiveness of HyCAM in complex multi-task scenarios, we construct a comprehensive benchmark dataset comprising five distinct domains: Auto CoT (logical reasoning), iCliniq (medical QA), Dolly 2.0 (general instruction-following), CodeAlpaca (code generation), and WebGPT (information retrieval QA).
Detailed information, including statistics, domains, and sources, is presented in Table \ref{tab:dataset}. Further, for robust evaluation across these datasets, we employ a 7:2:1 split for training, validation, and testing, and conduct five-fold cross-validation.

\subsubsection{\textbf{Backbone Models}}
We evaluate HyCAM across three state-of-the-art open-source LLM families to demonstrate its applicability:

\paratitle{LLaMA.} We utilize multiple versions from Meta AI's LLaMA series, which is known for strong performance in zero-shot and few-shot scenarios with large amounts of pre-training data.
Specifically, our experiments include Llama2-7B, Llama3-8B, Llama3.1-8B, and the smaller Llama3.2-1B/3B models.

\paratitle{Mistral.} Introduced by Mistral AI, 
Mistral features a compact model to achieve comparable performance to the mainstream methods via knowledge distillation. 
We use Mistral-7B-v0.3 as our base model.

\paratitle{Qwen.} 
Developed by Alibaba, Qwen focuses on efficient inference and robust performance across diverse tasks.
We adopt various sizes of Qwen2.5 series, ranging from 0.5B to up to 14B.

\subsubsection{\textbf{Baseline Methods}}
We evaluate HyCAM against several representative methods of task adaptation, detailed as follows:

\begin{itemize}[leftmargin=*]
    \item \textbf{Full Fine-Tuning} involves updating all parameters of the backbone LLM for adaptation to the target tasks.
    \item \textbf{LoRA}~\cite{hu2021lora} injects trainable low-rank adaptation matrices, allowing efficient adaptation with reduced trainable parameters.  
    \item \textbf{\multilora}~\cite{wang2023multilora} applies multiple LoRA adapters in parallel, enabling independent adaptation for different tasks.  
    \item \textbf{\moelora}~\cite{sun2025stronger} integrates Riemannian gradient rescaling with MoE-LoRA to preserve expressiveness while stabilizing optimization and accelerating convergence.
\end{itemize}

\subsubsection{\textbf{Evaluation Metrics}} To comprehensively evaluate the performance of methods across diverse tasks and domains, we adopt three commonly used evaluation metrics:

\noindent\textbf{PPL} (Perplexity) assesses the fluency and overall language generation quality. Lower PPL scores indicate better performance.\\
\textbf{BLEU-4} (Bilingual Evaluation Understudy)~\cite{papineni2002bleu} measures the n-gram overlap between generated and label texts, particularly relevant for tasks like code generation. Higher BLEU-4 scores are better. \\
\textbf{ROUGE-L} (Recall-Oriented Understudy for Gisting Evaluation - Longest Common Subsequence)~\cite{lin2004rouge} captures content overlap and semantic similarity, especially for longer outputs like summaries. Higher ROUGE-L scores indicate better performance.

\subsubsection{\textbf{Implementation Details}}
All our experiments are implemented using PyTorch, and we leverage DeepSpeed for efficient training with BFloat16 mixed-precision.
For LoRA-based methods, the LoRA rank ($r$) is set to 64 and is applied to all linear layers of LLM.
For methods involving multiple specialized modules, the number of modules ($N_s$) is set to 5.
The maximum token length is set to 1,200 to accommodate long-text tasks.
For HyCAM-specific hyperparameters, the Gumbel-Softmax temperature $\tau$ is set to 0.5, and the load-balancing loss coefficient $\lambda_{\text{balance}}$ is set to 0.1.
For training, we use the AdamW optimizer with a learning rate of 2e-5, a cosine decay learning rate scheduler, and early stopping based on validation loss to prevent overfitting.
\subsection{Overall Performance (RQ1)}
To answer \textbf{RQ1}, we evaluate the overall multi-task adaptation performance of HyCAM against the baseline methods.
The experiment is conducted across different backbone LLMs of comparable scale (7B/8B parameters) with our comprehensive multi-task datasets.
The main results, summarized as the average performance across all tasks, are presented in Table~\ref{tab:exp1}.

The experimental results demonstrate the superior overall performance of HyCAM.
On average, HyCAM achieves a 3.65\% relative improvement across all metrics and backbone LLMs compared to the best baseline, with statistically significant ($p$ < 0.05, indicated by an asterisk~($\ast$) in the table). 

Compared to Full Fine-Tuning, HyCAM achieves excellent performance while only updating a small fraction of the parameters, underscoring its advantage in computational efficiency.
Compared with the single-task PEFT method LoRA, HyCAM shows substantial gains. While LoRA provides a parameter-efficient alternative to full fine-tuning, its capacity can be limited in complex multi-task scenarios. 
HyCAM, with its CAM mechanism and hybrid architecture design, is better able to handle the diverse demands of multiple tasks simultaneously, thus achieving better adaptation performance.

Furthermore, HyCAM outperforms other multi-task approaches, including \multilora and \moelora.
While \multilora allows for parallel task adaptations, it may not facilitate effective knowledge sharing between tasks. 
\moelora, despite its gradient rescaling mechanism, may struggle with optimal expert utilization.
HyCAM's integration of Shared CAM and Specialized CAM, with a dynamic routing strategy, provides a more effective framework for both knowledge sharing and specialized adaptation.

\vspace{-5px}
\subsection{Scalability Analysis (RQ2)}
To address \textbf{RQ2} and understand how the effectiveness of HyCAM scales with the size of backbone model, we evaluated its performance across a range of model size within two LLM families: Qwen2.5 (from 0.5B to 14B parameters) and Llama3.2 (1B and 3B parameters), as shown in Table \ref{tab:expdiffqwen}, \ref{tab:expsmallllama}. 
This analysis aims to determine if the advantages are consistent across different model sizes and whether they become more or less significant with increasing size. 

The results indicate that HyCAM consistently outperforms other PEFT-based methods across all tested model sizes within both the Qwen and Llama families. 
While Full Fine-Tuning remains a strong baseline, HyCAM consistently offers a more parameter-efficient solution with competitive, and often superior, performance.
Moreover, a key observation is that the relative advantage of HyCAM often becomes more obvious as the model size increases. This suggests that larger models, with their greater capacity and more extensive general knowledge, benefit even more from HyCAM's ability to dynamically modulate attention and integrate shared and specialized knowledge effectively. 

\vspace{-5px}
\subsection{Task-Specific Analysis (RQ3)}
To address RQ3, we analyze the performance on individual tasks within our multi-task dataset.
We break down the results obtained with the Llama2-7B backbone model, as presented in Table~\ref{tab:crosstaskresult}.
The results indicate that HyCAM generally achieves competitive performance across individual tasks.
It is worth noting that performance varies considerably across tasks, due to differences in inherent complexity and task characteristics.
In particular, the results on the iClinq and Dolly datasets are significantly lower than those of others.
This observation highlights the importance of considering task complexity when designing multi-task learning frameworks for real-world applications.

\begin{figure}[t]
    \centering
    \includegraphics[width=0.9\linewidth]{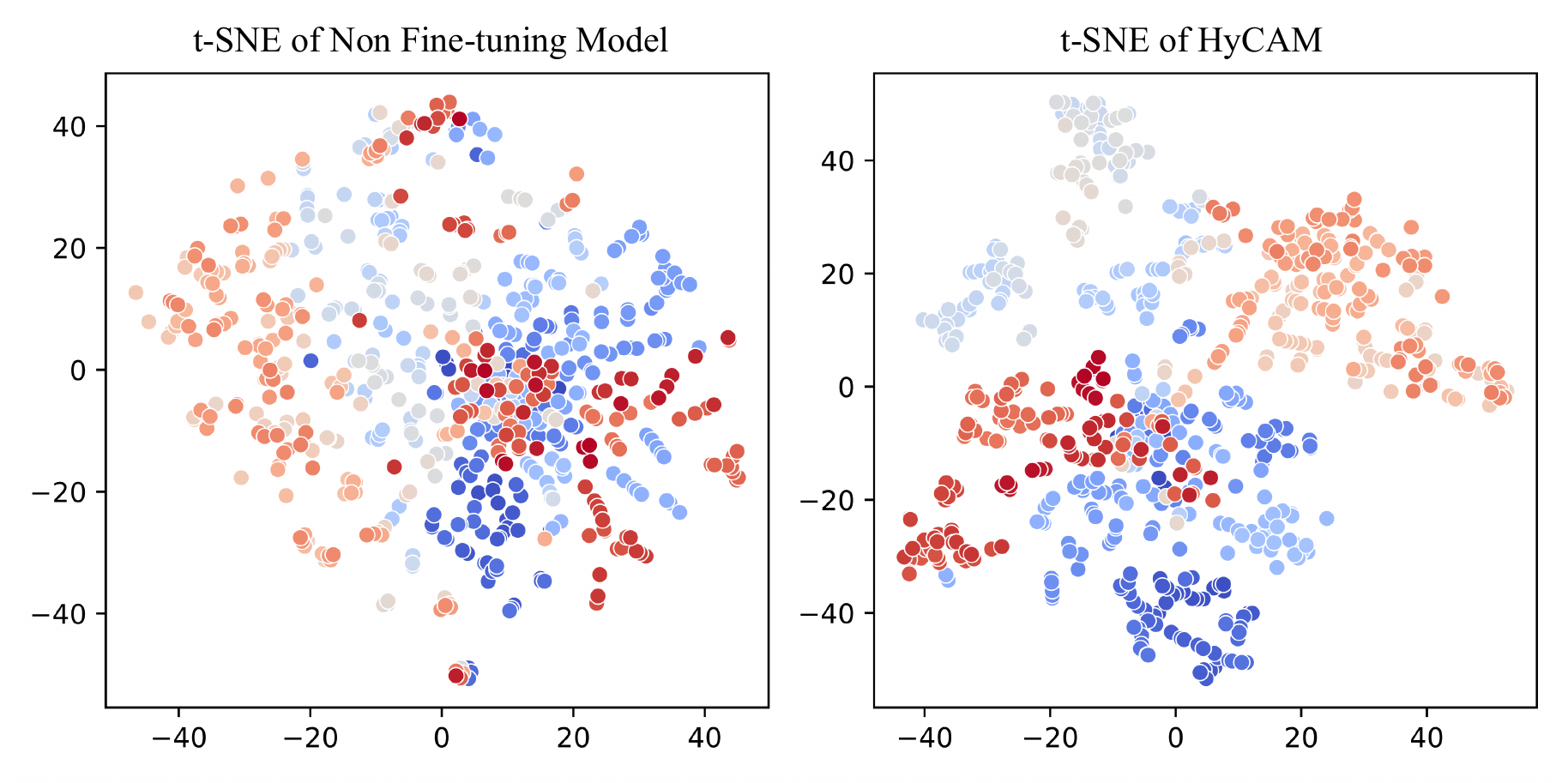}
    \caption{Scatter plots of attention representations. Higher density indicates improved representation capacity of the attention module.}
    \label{fig:tSNE}
\end{figure}

\subsection{Ablation and Hyperparameter Analysis (RQ4)}
\begin{figure}[tb]
  \centering
  \begin{minipage}[t]{.48\columnwidth}
    \vspace{2pt}
    \centering
    \includegraphics[width=0.8\linewidth]{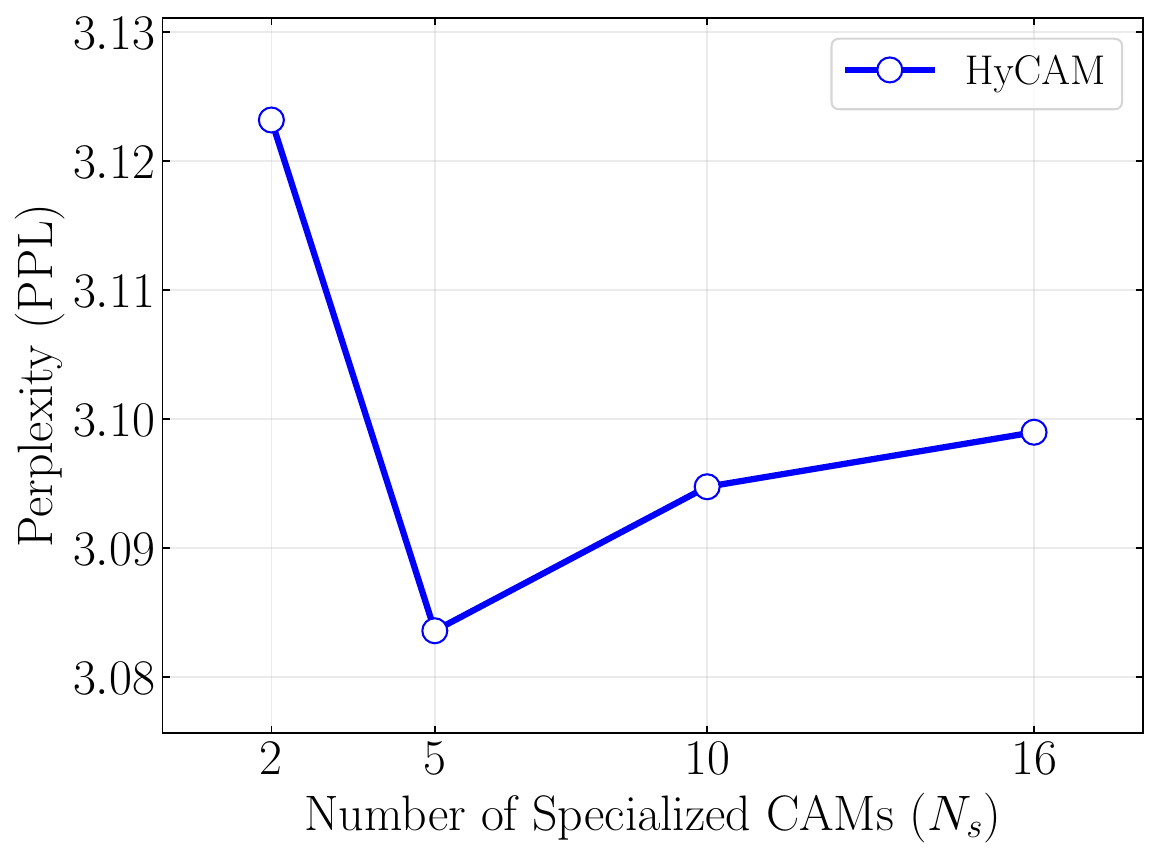}
    \captionsetup{font=small}
    \captionof{figure}{Impact of Number of Specialized CAMs.}
    \label{fig:cam_sens}
  \end{minipage}
  \hfill
  \begin{minipage}[t]{.48\columnwidth}
    \centering
    \captionsetup{font=small}
    \captionof{table}{Ablation study of HyCAM on Llama 2 7B.}
    \vspace{-1pt}
      \renewcommand{\arraystretch}{1}
      \resizebox{.8\linewidth}{!}{
        \begin{tabular}{lc}
          \toprule
          Variant & PPL$\downarrow$\\
          \midrule
          Shared-CAM-Only   & 3.129\\
          HyCAM-FullSpec    & 3.102\\
          HyCAM-SpecOnly    & 3.216\\
          HyCAM-InversePEFT & 3.129\\
          \textbf{HyCAM}    & \textbf{3.081}\\
          \bottomrule
        \end{tabular}}
    \label{tab:hycam_ablation}
  \end{minipage}
\end{figure}

For \textbf{RQ4}, we investigate the contributions of HyCAM components and their sensitivity to key hyperparameters. 
All studies here are conducted on the Llama2-7B model with PPL metric.
First, we performed ablation studies by evaluating several variants of HyCAM: 
\begin{itemize}[leftmargin=*, topsep=0pt]
\item \textbf{Shared-CAM-Only}: Only the single, shared full-parameter CAM module, removing all specialized CAMs and the routing mechanism, to assess the baseline impact of CAM.
\item \textbf{HyCAM-FullSpec}: Both the shared CAM and all Specialized CAMs are implemented with full parameters, to evaluate the benefit of using PEFT for the specialized components.
\item \textbf{HyCAM-SpecOnly}: Removes the shared, full-parameter CAM, relying exclusively on the ensemble of PEFT-based Specialized CAMs managed by the dynamic router and load-balancing loss.
\item \textbf{HyCAM-InversePEFT}: A special configuration where the Shared CAM module is implemented using PEFT, while the Specialized CAMs are full-parameter CAM modules, to validate our architectural design for parameter allocation.
\end{itemize}
Overall, these experiments demonstrate that each component and design choice in HyCAM contributes positively to its overall performance, as shown in Table~\ref{tab:hycam_ablation}.
We further analyzed HyCAM's sensitivity to the number of Specialized CAM modules ($N_s$), as illustrated in Figure~\ref{fig:cam_sens}.

\subsection{Qualitative Evaluation (RQ5)}
\begin{figure}[tb]
  \centering
  \begin{minipage}[t]{.45\columnwidth}
    \vspace{2pt}
    \centering
      \includegraphics[width=\linewidth]{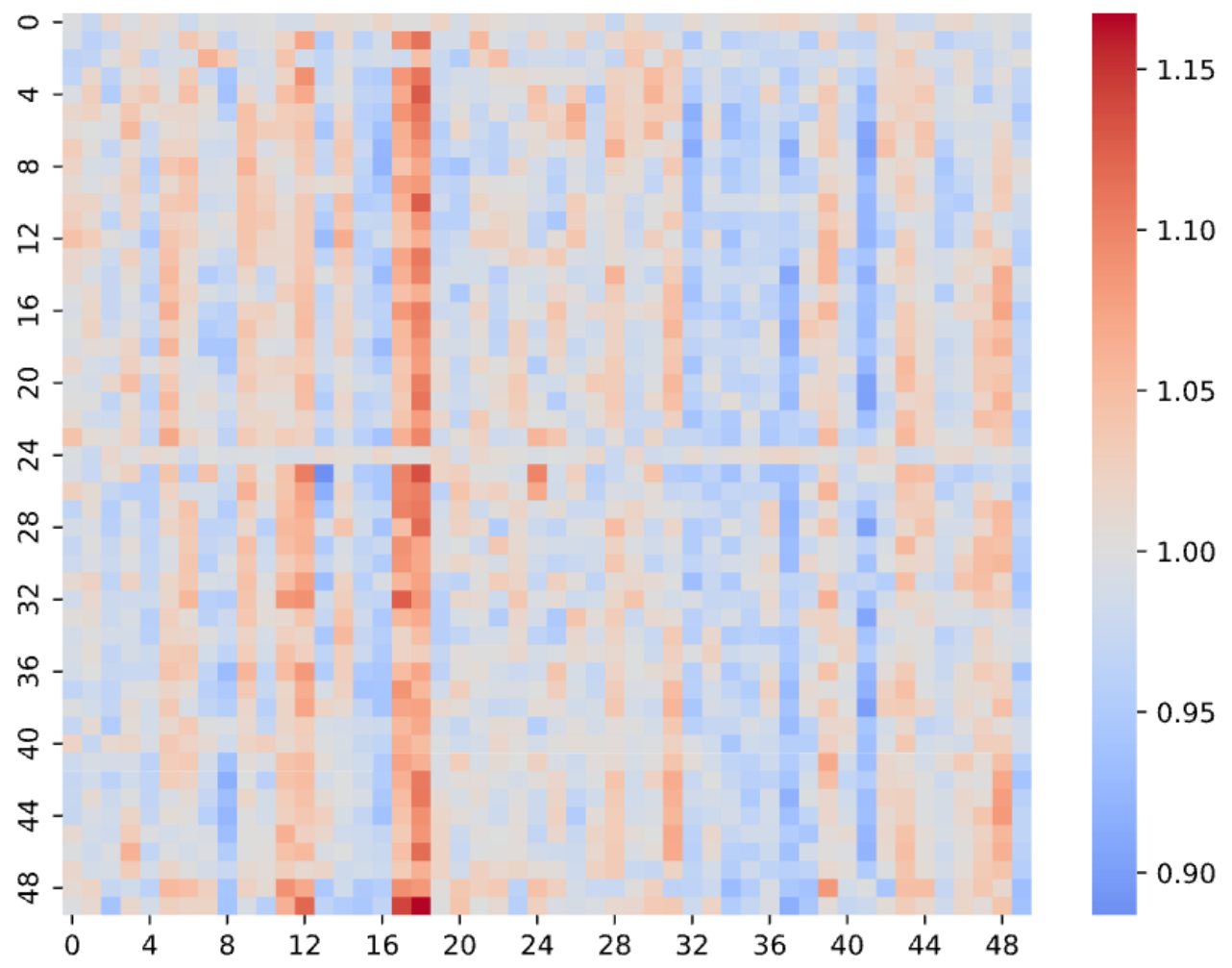}
    \captionsetup{font=small}
    \captionof{figure}{The Weight matrix of HyCAM.}
    \label{fig:ht}
  \end{minipage}
  \hfill
  \begin{minipage}[t]{.54\columnwidth}
    \vspace{12pt}
    \centering
      \includegraphics[width=\linewidth]{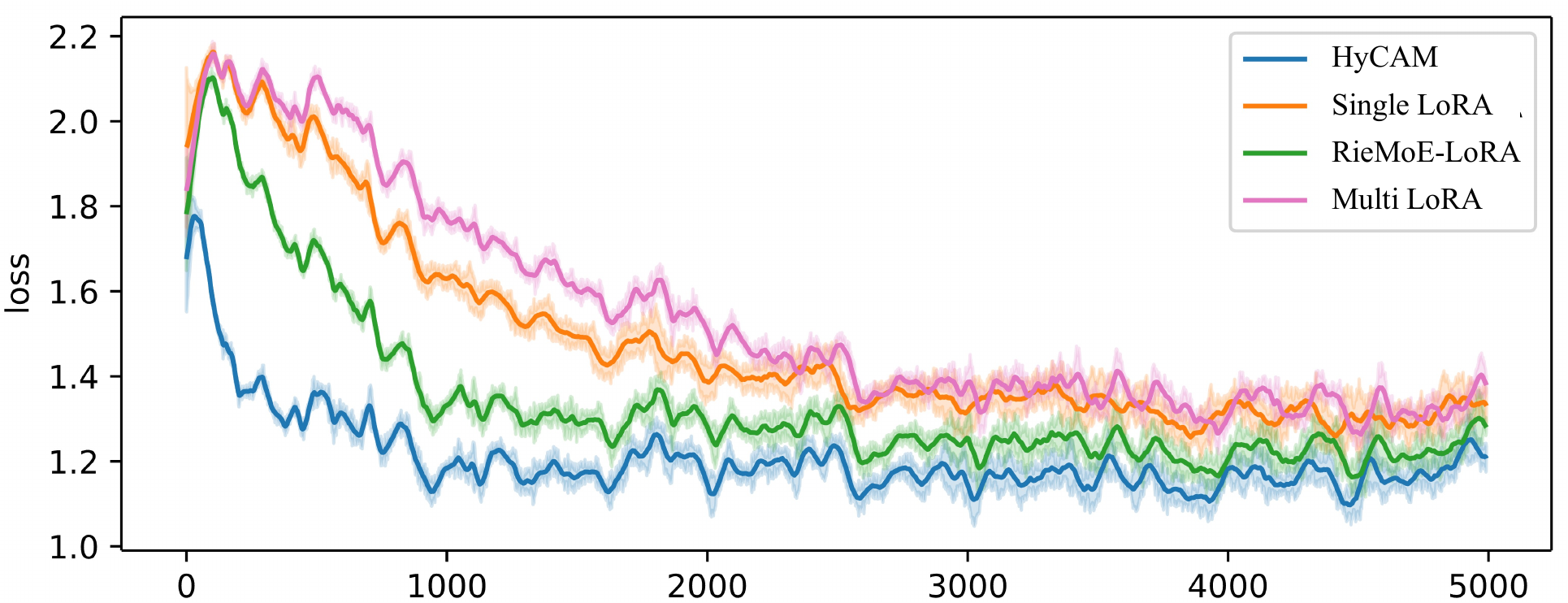}
    \captionsetup{font=small}
    \captionof{figure}{Comparison of the training processes across different methods.}
    \label{fig:loss}
  \end{minipage}
\end{figure}

To answer \textbf{RQ5} and get in-depth insights into the inner mechanisms of HyCAM, we employ several visualization techniques: 
\begin{itemize}[leftmargin=*, topsep=0pt]
\item \textbf{Enhanced Representational Coherence:} We utilize t-SNE to visualize the representations of the value matrix ($V$) within the self-attention module after applying our HyCAM. 
The vertical axis represents each token, and the horizontal axis indicates the degrees to be enhanced.
As shown in Figure~\ref{fig:tSNE}, CAM leads to more coherent clusters in the learned representations compared to a non-modulated baseline. This suggests an improved capacity to form meaningful representations.
\item \textbf{Feature Selective Modulation:} To understand how CAM impacts features, we visualize the modulation weight matrix generated by the HyCAM mechanism. As in Figure~\ref{fig:ht}, these weights demonstrate selective amplification of features relevant to the input context, while other features are attenuated. This highlights CAM's ability to perform context-dependent adjustments.
\item \textbf{Accelerated Convergence:} Figure~\ref{fig:loss} presents a comparison of training loss curves for HyCAM against baseline methods. These curves illustrate that HyCAM achieves lower loss more rapidly and stably, indicating efficient learning ability.
\end{itemize}

\section{Related Work}

\subsection{Parameter-Efficient Fine-Tuning}
To improve the efficiency of LLMs and make them more practical for real-world applications~\cite{liu2024moe}, various methods, such as pruning~\cite{wang2025put}, compression~\cite{wang2023large}, and PEFT methods have been proposed.
The primary goal of PEFT methods is to adapt LLM to specific tasks by updating only a small fraction of parameters, thereby preserving pre-trained knowledge and significantly reducing computational and memory cost~\cite{han2024parameter, fu2025sliding}. 
Conventional PEFT strategies include additive methods like adapters, selective fine-tuning, and reparameterization techniques such as low-rank adaptation.
Adapters, such as AdapterFusion~\cite{pfeiffer2020adapterfusion}, involve inserting lightweight, task-specific modules into the layers of the pre-trained model. 
Selective fine-tuning approaches, such as BitFit~\cite{zaken2021bitfit}, demonstrate that even minimal parameter adjustments can be effective for many tasks.
Reparameterization methods, such as Low-rank adaptation (LoRA)~\cite{hu2021lora} and its variants like AdaLoRA, DoRA, and MetaLoRA~\cite{zhang2023adalora, liu2024dora, wang2025metalora}, apply rank decomposition into target layers. LoRA operates on the principle that updates to the weight matrices during adaptation possess a low intrinsic rank.
Despite their success in reducing computational demands, existing PEFT methods often perform suboptimally in complex multi-task scenarios. A key challenge is balancing knowledge retention with task-specific specialization across multiple, potentially conflicting, objectives. Many PEFT approaches may exhibit limited generalization and representational capacity across diverse tasks, or suffer from potential interference when adapted to multiple objectives simultaneously. 

\subsection{Multi-task Adaptation Methods}
Multi-task learning (MTL) is fundamental across many real-world domains, including recommendation systems~\cite{liu2023multi, liu2025multi}, environmental prediction~\cite{hettige2024airphynet, han2025bridging, ji2023multi}, and heterogeneous time-series analysis~\cite{du2021gan,wang2023wavelet}.
Within LLMs, MTL aims to enhance generalization by sharing knowledge across tasks~\cite{guo2024large, zhao2023survey, wang2024llm4msr}, but a key challenge lies in balancing shared knowledge with task-specific specialization~\cite{gao2024higher}.
Several strategies have been explored for MTL in LLMs. A common paradigm involves hard parameter sharing, such as T5~\cite{raffel2020exploring}. 
PEFT techniques have been adapted for MTL. For instance, methods like Multi-LoRA~\cite{wang2023multilora} and adapter-based methods~\cite{pfeiffer2020mad} use small trainable modules into the frozen pre-trained model.
Another type of multi-task method is the Mixture-of-Experts (MoE) framework, such as Switch Transformers~\cite{fedus2022switch} and GShard~\cite{lepikhin2020gshard}, which employ a routing mechanism that activates a subset of experts for each input. MoE-LoRA~\cite{luo2024moelora} introduces Layer-wise Expert Allocation (MoLA), which assigns experts at different Transformer layers for better adaptability.
While powerful, MoE-based approaches often face challenges such as expert load imbalance, which influences training efficiency.
Our proposed HyCAM framework offers a novel approach to these challenges by combining CAM with a hybrid strategy for knowledge sharing and specialization, enabling efficient knowledge sharing while preserving task-specific adaptations.

\section{Conclusion}
In this work, we propose HyCAM, a novel framework designed to enhance multi-task adaptation in Large Language Models by integrating our core CAM mechanism within a hybrid architecture.
HyCAM effectively balances generalization and specialization by employing a shared, full-parameter CAM module for broad knowledge retention and multiple specialized, lightweight CAM modules for fine-grained, task-specific feature enhancement. Our dynamic soft-routing strategy, with a load-balancing loss, ensures adaptive knowledge fusion and efficient utilization of these specialized components. 
Extensive experiments show that HyCAM significantly outperforms existing approaches, demonstrating its ability for efficient adaptation and preserving pre-trained general knowledge.

\begin{acks}
Jingyuan Wang's work was partially supported by the National Natural Science Foundation of China (No. 72171013, 72222022, 72242101), and the Fundamental Research Funds for the Central Universities (JKF-2025017226182).
This research was partially supported by Hong Kong Research Grants Council's Research Impact Fund (No.R1015-23), Collaborative Research Fund (No.C1043-24GF), General Research Fund (No.11218325), Institute of Digital Medicine of City University of Hong Kong (No.9229503), National Natural Science Foundation of China (No.62502404), Huawei (Huawei Innovation Research Program), Tencent (CCF-Tencent Open Fund, Tencent Rhino-Bird Focused Research Program), Alibaba (CCF-Alimama Tech Kangaroo Fund No. 2024002), Ant Group (CCF-Ant Research Fund), Kuaishou, Didi (CCF-Didi Gaia Scholars Research Fund), and Bytedance.
\end{acks}

\section{GenAI Usage Disclosure}
This research, including the proposed Contextual Attention Modulation (CAM) and Hybrid Contextual Attention Modulation (HyCAM) framework, the experiment design and the analysis of experimental results, are the original contributions of the authors. 

Our work involves efficient multi-task adaptation of LLMs, and we need to use GenAI tools and models during the research process, as follows:

1.  Use in experiments: Our study involved evaluating the proposed HyCAM framework for multi-task adaptation in LLMs.
To this end, GenAI models, specifically variants from the Llama, Mistral, and Qwen families, are employed as backbone models.
These pre-trained LLMs were subjected to fine-tuning using our proposed method and baseline approaches, as described in the experimental sections.

2.  Use in Data Origins: Certain datasets utilized in our multi-task benchmark were constructed with the involvement of GenAI. 
Datasets such as Auto CoT, CodeAlpaca, and WebGPT may incorporate machine generated data, but no additional synthetic data were generated for this study.

The authors declare that they comply with the ACM policy on the usage of GenAI.

\bibliographystyle{ACM-Reference-Format}
\balance
\bibliography{ref}


\begin{thebibliography}{68}


\ifx \showCODEN    \undefined \def \showCODEN     #1{\unskip}     \fi
\ifx \showISBNx    \undefined \def \showISBNx     #1{\unskip}     \fi
\ifx \showISBNxiii \undefined \def \showISBNxiii  #1{\unskip}     \fi
\ifx \showISSN     \undefined \def \showISSN      #1{\unskip}     \fi
\ifx \showLCCN     \undefined \def \showLCCN      #1{\unskip}     \fi
\ifx \shownote     \undefined \def \shownote      #1{#1}          \fi
\ifx \showarticletitle \undefined \def \showarticletitle #1{#1}   \fi
\ifx \showURL      \undefined \def \showURL       {\relax}        \fi
\providecommand\bibfield[2]{#2}
\providecommand\bibinfo[2]{#2}
\providecommand\natexlab[1]{#1}
\providecommand\showeprint[2][]{arXiv:#2}

\bibitem[Achiam et~al\mbox{.}(2023)]%
        {achiam2023gpt}
\bibfield{author}{\bibinfo{person}{Josh Achiam}, \bibinfo{person}{Steven Adler}, \bibinfo{person}{Sandhini Agarwal}, \bibinfo{person}{Lama Ahmad}, \bibinfo{person}{Ilge Akkaya}, \bibinfo{person}{Florencia~Leoni Aleman}, \bibinfo{person}{Diogo Almeida}, \bibinfo{person}{Janko Altenschmidt}, \bibinfo{person}{Sam Altman}, \bibinfo{person}{Shyamal Anadkat}, {et~al\mbox{.}}} \bibinfo{year}{2023}\natexlab{}.
\newblock \showarticletitle{Gpt-4 technical report}.
\newblock \bibinfo{journal}{\emph{arXiv preprint arXiv:2303.08774}} (\bibinfo{year}{2023}).
\newblock


\bibitem[Ba et~al\mbox{.}(2016)]%
        {ba2016layer}
\bibfield{author}{\bibinfo{person}{Jimmy~Lei Ba}, \bibinfo{person}{Jamie~Ryan Kiros}, {and} \bibinfo{person}{Geoffrey~E Hinton}.} \bibinfo{year}{2016}\natexlab{}.
\newblock \showarticletitle{Layer normalization}.
\newblock \bibinfo{journal}{\emph{arXiv preprint arXiv:1607.06450}} (\bibinfo{year}{2016}).
\newblock


\bibitem[Bommasani et~al\mbox{.}(2021)]%
        {bommasani2021opportunities}
\bibfield{author}{\bibinfo{person}{Rishi Bommasani}, \bibinfo{person}{Drew~A Hudson}, \bibinfo{person}{Ehsan Adeli}, \bibinfo{person}{Russ Altman}, \bibinfo{person}{Simran Arora}, \bibinfo{person}{Sydney von Arx}, \bibinfo{person}{Michael~S Bernstein}, \bibinfo{person}{Jeannette Bohg}, \bibinfo{person}{Antoine Bosselut}, \bibinfo{person}{Emma Brunskill}, {et~al\mbox{.}}} \bibinfo{year}{2021}\natexlab{}.
\newblock \showarticletitle{On the opportunities and risks of foundation models}.
\newblock \bibinfo{journal}{\emph{arXiv preprint arXiv:2108.07258}} (\bibinfo{year}{2021}).
\newblock


\bibitem[Brynjolfsson et~al\mbox{.}(2025)]%
        {brynjolfsson2025generative}
\bibfield{author}{\bibinfo{person}{Erik Brynjolfsson}, \bibinfo{person}{Danielle Li}, {and} \bibinfo{person}{Lindsey Raymond}.} \bibinfo{year}{2025}\natexlab{}.
\newblock \showarticletitle{Generative AI at work}.
\newblock \bibinfo{journal}{\emph{The Quarterly Journal of Economics}} (\bibinfo{year}{2025}), \bibinfo{pages}{qjae044}.
\newblock


\bibitem[Cai et~al\mbox{.}(2024)]%
        {cai2024survey}
\bibfield{author}{\bibinfo{person}{Weilin Cai}, \bibinfo{person}{Juyong Jiang}, \bibinfo{person}{Fan Wang}, \bibinfo{person}{Jing Tang}, \bibinfo{person}{Sunghun Kim}, {and} \bibinfo{person}{Jiayi Huang}.} \bibinfo{year}{2024}\natexlab{}.
\newblock \showarticletitle{A survey on mixture of experts}.
\newblock \bibinfo{journal}{\emph{arXiv preprint arXiv:2407.06204}} (\bibinfo{year}{2024}).
\newblock


\bibitem[Chaudhary(2023)]%
        {codealpaca}
\bibfield{author}{\bibinfo{person}{Sahil Chaudhary}.} \bibinfo{year}{2023}\natexlab{}.
\newblock \bibinfo{title}{Code Alpaca: An Instruction-following LLaMA model for code generation}.
\newblock \bibinfo{howpublished}{\url{https://github.com/sahil280114/codealpaca}}.
\newblock


\bibitem[Cheng et~al\mbox{.}(2025)]%
        {cheng2025poi}
\bibfield{author}{\bibinfo{person}{Jiawei Cheng}, \bibinfo{person}{Jingyuan Wang}, \bibinfo{person}{Yichuan Zhang}, \bibinfo{person}{Jiahao Ji}, \bibinfo{person}{Yuanshao Zhu}, \bibinfo{person}{Zhibo Zhang}, {and} \bibinfo{person}{Xiangyu Zhao}.} \bibinfo{year}{2025}\natexlab{}.
\newblock \showarticletitle{Poi-enhancer: An llm-based semantic enhancement framework for poi representation learning}. In \bibinfo{booktitle}{\emph{Proceedings of the AAAI conference on artificial intelligence}}, Vol.~\bibinfo{volume}{39}. \bibinfo{pages}{11509--11517}.
\newblock


\bibitem[Conover et~al\mbox{.}(2023)]%
        {DatabricksBlog2023DollyV2}
\bibfield{author}{\bibinfo{person}{Mike Conover}, \bibinfo{person}{Matt Hayes}, \bibinfo{person}{Ankit Mathur}, \bibinfo{person}{Jianwei Xie}, \bibinfo{person}{Jun Wan}, \bibinfo{person}{Sam Shah}, \bibinfo{person}{Ali Ghodsi}, \bibinfo{person}{Patrick Wendell}, \bibinfo{person}{Matei Zaharia}, {and} \bibinfo{person}{Reynold Xin}.} \bibinfo{year}{2023}\natexlab{}.
\newblock \bibinfo{booktitle}{\emph{Free Dolly: Introducing the World's First Truly Open Instruction-Tuned LLM}}.
\newblock
\urldef\tempurl%
\url{https://www.databricks.com/blog/2023/04/12/dolly-first-open-commercially-viable-instruction-tuned-llm}
\showURL{%
\tempurl}


\bibitem[Du et~al\mbox{.}(2021)]%
        {du2021gan}
\bibfield{author}{\bibinfo{person}{Bowen Du}, \bibinfo{person}{Xuanxuan Sun}, \bibinfo{person}{Junchen Ye}, \bibinfo{person}{Ke Cheng}, \bibinfo{person}{Jingyuan Wang}, {and} \bibinfo{person}{Leilei Sun}.} \bibinfo{year}{2021}\natexlab{}.
\newblock \showarticletitle{GAN-based anomaly detection for multivariate time series using polluted training set}.
\newblock \bibinfo{journal}{\emph{IEEE Transactions on Knowledge and Data Engineering}} \bibinfo{volume}{35}, \bibinfo{number}{12} (\bibinfo{year}{2021}), \bibinfo{pages}{12208--12219}.
\newblock


\bibitem[Elfwing et~al\mbox{.}(2018)]%
        {elfwing2018sigmoid}
\bibfield{author}{\bibinfo{person}{Stefan Elfwing}, \bibinfo{person}{Eiji Uchibe}, {and} \bibinfo{person}{Kenji Doya}.} \bibinfo{year}{2018}\natexlab{}.
\newblock \showarticletitle{Sigmoid-weighted linear units for neural network function approximation in reinforcement learning}.
\newblock \bibinfo{journal}{\emph{Neural networks}}  \bibinfo{volume}{107} (\bibinfo{year}{2018}), \bibinfo{pages}{3--11}.
\newblock


\bibitem[Fedus et~al\mbox{.}(2022)]%
        {fedus2022switch}
\bibfield{author}{\bibinfo{person}{William Fedus}, \bibinfo{person}{Barret Zoph}, {and} \bibinfo{person}{Noam Shazeer}.} \bibinfo{year}{2022}\natexlab{}.
\newblock \showarticletitle{Switch transformers: Scaling to trillion parameter models with simple and efficient sparsity}.
\newblock \bibinfo{journal}{\emph{Journal of Machine Learning Research}} \bibinfo{volume}{23}, \bibinfo{number}{120} (\bibinfo{year}{2022}), \bibinfo{pages}{1--39}.
\newblock


\bibitem[Fu et~al\mbox{.}(2025a)]%
        {fu2025sliding}
\bibfield{author}{\bibinfo{person}{Zichuan Fu}, \bibinfo{person}{Wentao Song}, \bibinfo{person}{Yejing Wang}, \bibinfo{person}{Xian Wu}, \bibinfo{person}{Yefeng Zheng}, \bibinfo{person}{Yingying Zhang}, \bibinfo{person}{Derong Xu}, \bibinfo{person}{Xuetao Wei}, \bibinfo{person}{Tong Xu}, {and} \bibinfo{person}{Xiangyu Zhao}.} \bibinfo{year}{2025}\natexlab{a}.
\newblock \showarticletitle{Sliding Window Attention Training for Efficient Large Language Models}.
\newblock \bibinfo{journal}{\emph{arXiv preprint arXiv:2502.18845}} (\bibinfo{year}{2025}).
\newblock


\bibitem[Fu et~al\mbox{.}(2025b)]%
        {fu2025training}
\bibfield{author}{\bibinfo{person}{Zichuan Fu}, \bibinfo{person}{Xian Wu}, \bibinfo{person}{Yejing Wang}, \bibinfo{person}{Wanyu Wang}, \bibinfo{person}{Shanshan Ye}, \bibinfo{person}{Hongzhi Yin}, \bibinfo{person}{Yi Chang}, \bibinfo{person}{Yefeng Zheng}, {and} \bibinfo{person}{Xiangyu Zhao}.} \bibinfo{year}{2025}\natexlab{b}.
\newblock \showarticletitle{Training-free LLM Merging for Multi-task Learning}.
\newblock \bibinfo{journal}{\emph{arXiv preprint arXiv:2506.12379}} (\bibinfo{year}{2025}).
\newblock


\bibitem[Gao et~al\mbox{.}(2024)]%
        {gao2024higher}
\bibfield{author}{\bibinfo{person}{Chongyang Gao}, \bibinfo{person}{Kezhen Chen}, \bibinfo{person}{Jinmeng Rao}, \bibinfo{person}{Baochen Sun}, \bibinfo{person}{Ruibo Liu}, \bibinfo{person}{Daiyi Peng}, \bibinfo{person}{Yawen Zhang}, \bibinfo{person}{Xiaoyuan Guo}, \bibinfo{person}{Jie Yang}, {and} \bibinfo{person}{VS Subrahmanian}.} \bibinfo{year}{2024}\natexlab{}.
\newblock \showarticletitle{Higher layers need more lora experts}.
\newblock \bibinfo{journal}{\emph{arXiv preprint arXiv:2402.08562}} (\bibinfo{year}{2024}).
\newblock


\bibitem[Geva et~al\mbox{.}(2021)]%
        {geva2021transformer}
\bibfield{author}{\bibinfo{person}{Mor Geva}, \bibinfo{person}{Roei Schuster}, \bibinfo{person}{Jonathan Berant}, {and} \bibinfo{person}{Omer Levy}.} \bibinfo{year}{2021}\natexlab{}.
\newblock \showarticletitle{Transformer Feed-Forward Layers Are Key-Value Memories}. In \bibinfo{booktitle}{\emph{Proceedings of the 2021 Conference on Empirical Methods in Natural Language Processing}}. \bibinfo{pages}{5484--5495}.
\newblock


\bibitem[Guo et~al\mbox{.}(2025)]%
        {guo2025nlora}
\bibfield{author}{\bibinfo{person}{Chenlu Guo}, \bibinfo{person}{Yuan Wu}, {and} \bibinfo{person}{Yi Chang}.} \bibinfo{year}{2025}\natexlab{}.
\newblock \showarticletitle{NLoRA: Nystr$\backslash$" om-Initiated Low-Rank Adaptation for Large Language Models}.
\newblock \bibinfo{journal}{\emph{arXiv preprint arXiv:2502.14482}} (\bibinfo{year}{2025}).
\newblock


\bibitem[Guo et~al\mbox{.}(2024)]%
        {guo2024large}
\bibfield{author}{\bibinfo{person}{Taicheng Guo}, \bibinfo{person}{Xiuying Chen}, \bibinfo{person}{Yaqi Wang}, \bibinfo{person}{Ruidi Chang}, \bibinfo{person}{Shichao Pei}, \bibinfo{person}{Nitesh~V Chawla}, \bibinfo{person}{Olaf Wiest}, {and} \bibinfo{person}{Xiangliang Zhang}.} \bibinfo{year}{2024}\natexlab{}.
\newblock \showarticletitle{Large language model based multi-agents: A survey of progress and challenges}.
\newblock \bibinfo{journal}{\emph{arXiv preprint arXiv:2402.01680}} (\bibinfo{year}{2024}).
\newblock


\bibitem[Han et~al\mbox{.}(2025)]%
        {han2025bridging}
\bibfield{author}{\bibinfo{person}{Chengkai Han}, \bibinfo{person}{Jingyuan Wang}, \bibinfo{person}{Yongyao Wang}, \bibinfo{person}{Xie Yu}, \bibinfo{person}{Hao Lin}, \bibinfo{person}{Chao Li}, {and} \bibinfo{person}{Junjie Wu}.} \bibinfo{year}{2025}\natexlab{}.
\newblock \showarticletitle{Bridging traffic state and trajectory for dynamic road network and trajectory representation learning}. In \bibinfo{booktitle}{\emph{Proceedings of the AAAI Conference on Artificial Intelligence}}, Vol.~\bibinfo{volume}{39}. \bibinfo{pages}{11763--11771}.
\newblock


\bibitem[Han et~al\mbox{.}(2024)]%
        {han2024parameter}
\bibfield{author}{\bibinfo{person}{Zeyu Han}, \bibinfo{person}{Chao Gao}, \bibinfo{person}{Jinyang Liu}, \bibinfo{person}{Jeff Zhang}, {and} \bibinfo{person}{Sai~Qian Zhang}.} \bibinfo{year}{2024}\natexlab{}.
\newblock \showarticletitle{Parameter-efficient fine-tuning for large models: A comprehensive survey}.
\newblock \bibinfo{journal}{\emph{arXiv preprint arXiv:2403.14608}} (\bibinfo{year}{2024}).
\newblock


\bibitem[He et~al\mbox{.}(2015)]%
        {he2015delving}
\bibfield{author}{\bibinfo{person}{Kaiming He}, \bibinfo{person}{Xiangyu Zhang}, \bibinfo{person}{Shaoqing Ren}, {and} \bibinfo{person}{Jian Sun}.} \bibinfo{year}{2015}\natexlab{}.
\newblock \showarticletitle{Delving deep into rectifiers: Surpassing human-level performance on imagenet classification}. In \bibinfo{booktitle}{\emph{Proceedings of the IEEE international conference on computer vision}}. \bibinfo{pages}{1026--1034}.
\newblock


\bibitem[Hettige et~al\mbox{.}(2024)]%
        {hettige2024airphynet}
\bibfield{author}{\bibinfo{person}{Kethmi~Hirushini Hettige}, \bibinfo{person}{Jiahao Ji}, \bibinfo{person}{Shili Xiang}, \bibinfo{person}{Cheng Long}, \bibinfo{person}{Gao Cong}, {and} \bibinfo{person}{Jingyuan Wang}.} \bibinfo{year}{2024}\natexlab{}.
\newblock \showarticletitle{Airphynet: Harnessing physics-guided neural networks for air quality prediction}.
\newblock \bibinfo{journal}{\emph{arXiv preprint arXiv:2402.03784}} (\bibinfo{year}{2024}).
\newblock


\bibitem[Houlsby et~al\mbox{.}(2019)]%
        {houlsby2019parameter}
\bibfield{author}{\bibinfo{person}{Neil Houlsby}, \bibinfo{person}{Andrei Giurgiu}, \bibinfo{person}{Stanislaw Jastrzebski}, \bibinfo{person}{Bruna Morrone}, \bibinfo{person}{Quentin De~Laroussilhe}, \bibinfo{person}{Andrea Gesmundo}, \bibinfo{person}{Mona Attariyan}, {and} \bibinfo{person}{Sylvain Gelly}.} \bibinfo{year}{2019}\natexlab{}.
\newblock \showarticletitle{Parameter-efficient transfer learning for NLP}. In \bibinfo{booktitle}{\emph{International conference on machine learning}}. PMLR, \bibinfo{pages}{2790--2799}.
\newblock


\bibitem[Hu et~al\mbox{.}(2021)]%
        {hu2021lora}
\bibfield{author}{\bibinfo{person}{Edward~J Hu}, \bibinfo{person}{Yelong Shen}, \bibinfo{person}{Phillip Wallis}, \bibinfo{person}{Zeyuan Allen-Zhu}, \bibinfo{person}{Yuanzhi Li}, \bibinfo{person}{Shean Wang}, \bibinfo{person}{Lu Wang}, {and} \bibinfo{person}{Weizhu Chen}.} \bibinfo{year}{2021}\natexlab{}.
\newblock \showarticletitle{Lora: Low-rank adaptation of large language models}.
\newblock \bibinfo{journal}{\emph{arXiv preprint arXiv:2106.09685}} (\bibinfo{year}{2021}).
\newblock


\bibitem[Jang et~al\mbox{.}(2016)]%
        {jang2016categorical}
\bibfield{author}{\bibinfo{person}{Eric Jang}, \bibinfo{person}{Shixiang Gu}, {and} \bibinfo{person}{Ben Poole}.} \bibinfo{year}{2016}\natexlab{}.
\newblock \showarticletitle{Categorical reparameterization with gumbel-softmax}.
\newblock \bibinfo{journal}{\emph{arXiv preprint arXiv:1611.01144}} (\bibinfo{year}{2016}).
\newblock


\bibitem[Jaszczur et~al\mbox{.}(2021)]%
        {jaszczur2021sparse}
\bibfield{author}{\bibinfo{person}{Sebastian Jaszczur}, \bibinfo{person}{Aakanksha Chowdhery}, \bibinfo{person}{Afroz Mohiuddin}, \bibinfo{person}{Lukasz Kaiser}, \bibinfo{person}{Wojciech Gajewski}, \bibinfo{person}{Henryk Michalewski}, {and} \bibinfo{person}{Jonni Kanerva}.} \bibinfo{year}{2021}\natexlab{}.
\newblock \showarticletitle{Sparse is enough in scaling transformers}.
\newblock \bibinfo{journal}{\emph{Advances in Neural Information Processing Systems}}  \bibinfo{volume}{34} (\bibinfo{year}{2021}), \bibinfo{pages}{9895--9907}.
\newblock


\bibitem[Ji et~al\mbox{.}(2023)]%
        {ji2023multi}
\bibfield{author}{\bibinfo{person}{Jiahao Ji}, \bibinfo{person}{Jingyuan Wang}, \bibinfo{person}{Yu Mou}, {and} \bibinfo{person}{Cheng Long}.} \bibinfo{year}{2023}\natexlab{}.
\newblock \showarticletitle{Multi-factor spatio-temporal prediction based on graph decomposition learning}.
\newblock \bibinfo{journal}{\emph{arXiv preprint arXiv:2310.10374}} (\bibinfo{year}{2023}).
\newblock


\bibitem[Ji et~al\mbox{.}(2025)]%
        {ji2025seeing}
\bibfield{author}{\bibinfo{person}{Jiahao Ji}, \bibinfo{person}{Wentao Zhang}, \bibinfo{person}{Jingyuan Wang}, {and} \bibinfo{person}{Chao Huang}.} \bibinfo{year}{2025}\natexlab{}.
\newblock \showarticletitle{Seeing the unseen: Learning basis confounder representations for robust traffic prediction}.
\newblock  (\bibinfo{year}{2025}).
\newblock


\bibitem[Jin et~al\mbox{.}(2025)]%
        {jin2025massive}
\bibfield{author}{\bibinfo{person}{Mingyu Jin}, \bibinfo{person}{Kai Mei}, \bibinfo{person}{Wujiang Xu}, \bibinfo{person}{Mingjie Sun}, \bibinfo{person}{Ruixiang Tang}, \bibinfo{person}{Mengnan Du}, \bibinfo{person}{Zirui Liu}, {and} \bibinfo{person}{Yongfeng Zhang}.} \bibinfo{year}{2025}\natexlab{}.
\newblock \showarticletitle{Massive Values in Self-Attention Modules are the Key to Contextual Knowledge Understanding}.
\newblock \bibinfo{journal}{\emph{arXiv preprint arXiv:2502.01563}} (\bibinfo{year}{2025}).
\newblock


\bibitem[Lepikhin et~al\mbox{.}(2020)]%
        {lepikhin2020gshard}
\bibfield{author}{\bibinfo{person}{Dmitry Lepikhin}, \bibinfo{person}{HyoukJoong Lee}, \bibinfo{person}{Yuanzhong Xu}, \bibinfo{person}{Dehao Chen}, \bibinfo{person}{Orhan Firat}, \bibinfo{person}{Yanping Huang}, \bibinfo{person}{Maxim Krikun}, \bibinfo{person}{Noam Shazeer}, {and} \bibinfo{person}{Zhifeng Chen}.} \bibinfo{year}{2020}\natexlab{}.
\newblock \showarticletitle{Gshard: Scaling giant models with conditional computation and automatic sharding}.
\newblock \bibinfo{journal}{\emph{arXiv preprint arXiv:2006.16668}} (\bibinfo{year}{2020}).
\newblock


\bibitem[Lester et~al\mbox{.}(2021)]%
        {lester2021power}
\bibfield{author}{\bibinfo{person}{Brian Lester}, \bibinfo{person}{Rami Al-Rfou}, {and} \bibinfo{person}{Noah Constant}.} \bibinfo{year}{2021}\natexlab{}.
\newblock \showarticletitle{The power of scale for parameter-efficient prompt tuning}.
\newblock \bibinfo{journal}{\emph{arXiv preprint arXiv:2104.08691}} (\bibinfo{year}{2021}).
\newblock


\bibitem[Li et~al\mbox{.}(2023c)]%
        {li2023web}
\bibfield{author}{\bibinfo{person}{Junyi Li}, \bibinfo{person}{Tianyi Tang}, \bibinfo{person}{Wayne~Xin Zhao}, \bibinfo{person}{Jingyuan Wang}, \bibinfo{person}{Jian-Yun Nie}, {and} \bibinfo{person}{Ji-Rong Wen}.} \bibinfo{year}{2023}\natexlab{c}.
\newblock \showarticletitle{The web can be your oyster for improving large language models}.
\newblock \bibinfo{journal}{\emph{arXiv preprint arXiv:2305.10998}} (\bibinfo{year}{2023}).
\newblock


\bibitem[Li et~al\mbox{.}(2023a)]%
        {li2023e4srec}
\bibfield{author}{\bibinfo{person}{Xinhang Li}, \bibinfo{person}{Chong Chen}, \bibinfo{person}{Xiangyu Zhao}, \bibinfo{person}{Yong Zhang}, {and} \bibinfo{person}{Chunxiao Xing}.} \bibinfo{year}{2023}\natexlab{a}.
\newblock \showarticletitle{E4srec: An elegant effective efficient extensible solution of large language models for sequential recommendation}.
\newblock \bibinfo{journal}{\emph{arXiv preprint arXiv:2312.02443}} (\bibinfo{year}{2023}).
\newblock


\bibitem[Li and Liang(2021)]%
        {li2021prefix}
\bibfield{author}{\bibinfo{person}{Xiang~Lisa Li} {and} \bibinfo{person}{Percy Liang}.} \bibinfo{year}{2021}\natexlab{}.
\newblock \showarticletitle{Prefix-tuning: Optimizing continuous prompts for generation}.
\newblock \bibinfo{journal}{\emph{arXiv preprint arXiv:2101.00190}} (\bibinfo{year}{2021}).
\newblock


\bibitem[Li et~al\mbox{.}(2023b)]%
        {li2023chatdoctor}
\bibfield{author}{\bibinfo{person}{Yunxiang Li}, \bibinfo{person}{Zihan Li}, \bibinfo{person}{Kai Zhang}, \bibinfo{person}{Ruilong Dan}, \bibinfo{person}{Steve Jiang}, {and} \bibinfo{person}{You Zhang}.} \bibinfo{year}{2023}\natexlab{b}.
\newblock \showarticletitle{ChatDoctor: A Medical Chat Model Fine-Tuned on a Large Language Model Meta-AI (LLaMA) Using Medical Domain Knowledge}.
\newblock \bibinfo{journal}{\emph{Cureus}} \bibinfo{volume}{15}, \bibinfo{number}{6} (\bibinfo{year}{2023}).
\newblock


\bibitem[Lin(2004)]%
        {lin2004rouge}
\bibfield{author}{\bibinfo{person}{Chin-Yew Lin}.} \bibinfo{year}{2004}\natexlab{}.
\newblock \showarticletitle{Rouge: A package for automatic evaluation of summaries}. In \bibinfo{booktitle}{\emph{Text summarization branches out}}. \bibinfo{pages}{74--81}.
\newblock


\bibitem[Liu et~al\mbox{.}(2021)]%
        {liu2021conflict}
\bibfield{author}{\bibinfo{person}{Bo Liu}, \bibinfo{person}{Xingchao Liu}, \bibinfo{person}{Xiaojie Jin}, \bibinfo{person}{Peter Stone}, {and} \bibinfo{person}{Qiang Liu}.} \bibinfo{year}{2021}\natexlab{}.
\newblock \showarticletitle{Conflict-averse gradient descent for multi-task learning}.
\newblock \bibinfo{journal}{\emph{Advances in Neural Information Processing Systems}}  \bibinfo{volume}{34} (\bibinfo{year}{2021}), \bibinfo{pages}{18878--18890}.
\newblock


\bibitem[Liu et~al\mbox{.}(2025)]%
        {liu2025multi}
\bibfield{author}{\bibinfo{person}{Langming Liu}, \bibinfo{person}{Wanyu Wang}, \bibinfo{person}{Chi Zhang}, \bibinfo{person}{Bo Li}, \bibinfo{person}{Hongzhi Yin}, \bibinfo{person}{Xuetao Wei}, \bibinfo{person}{Wenbo Su}, \bibinfo{person}{Bo Zheng}, {and} \bibinfo{person}{Xiangyu Zhao}.} \bibinfo{year}{2025}\natexlab{}.
\newblock \showarticletitle{Multi-task Offline Reinforcement Learning for Online Advertising in Recommender Systems}. In \bibinfo{booktitle}{\emph{Proceedings of the 31st ACM SIGKDD Conference on Knowledge Discovery and Data Mining V. 2}}. \bibinfo{pages}{4635--4646}.
\newblock


\bibitem[Liu et~al\mbox{.}(2024b)]%
        {liu2024moe}
\bibfield{author}{\bibinfo{person}{Qidong Liu}, \bibinfo{person}{Xian Wu}, \bibinfo{person}{Xiangyu Zhao}, \bibinfo{person}{Yuanshao Zhu}, \bibinfo{person}{Derong Xu}, \bibinfo{person}{Feng Tian}, {and} \bibinfo{person}{Yefeng Zheng}.} \bibinfo{year}{2024}\natexlab{b}.
\newblock \showarticletitle{When moe meets llms: Parameter efficient fine-tuning for multi-task medical applications}. In \bibinfo{booktitle}{\emph{Proceedings of the 47th International ACM SIGIR Conference on Research and Development in Information Retrieval}}. \bibinfo{pages}{1104--1114}.
\newblock


\bibitem[Liu et~al\mbox{.}(2024a)]%
        {liu2024dora}
\bibfield{author}{\bibinfo{person}{Shih-Yang Liu}, \bibinfo{person}{Chien-Yi Wang}, \bibinfo{person}{Hongxu Yin}, \bibinfo{person}{Pavlo Molchanov}, \bibinfo{person}{Yu-Chiang~Frank Wang}, \bibinfo{person}{Kwang-Ting Cheng}, {and} \bibinfo{person}{Min-Hung Chen}.} \bibinfo{year}{2024}\natexlab{a}.
\newblock \showarticletitle{Dora: Weight-decomposed low-rank adaptation}.
\newblock \bibinfo{journal}{\emph{arXiv preprint arXiv:2402.09353}} (\bibinfo{year}{2024}).
\newblock


\bibitem[Liu et~al\mbox{.}(2023)]%
        {liu2023multi}
\bibfield{author}{\bibinfo{person}{Ziru Liu}, \bibinfo{person}{Jiejie Tian}, \bibinfo{person}{Qingpeng Cai}, \bibinfo{person}{Xiangyu Zhao}, \bibinfo{person}{Jingtong Gao}, \bibinfo{person}{Shuchang Liu}, \bibinfo{person}{Dayou Chen}, \bibinfo{person}{Tonghao He}, \bibinfo{person}{Dong Zheng}, \bibinfo{person}{Peng Jiang}, {et~al\mbox{.}}} \bibinfo{year}{2023}\natexlab{}.
\newblock \showarticletitle{Multi-task recommendations with reinforcement learning}. In \bibinfo{booktitle}{\emph{Proceedings of the ACM web conference 2023}}. \bibinfo{pages}{1273--1282}.
\newblock


\bibitem[Luo et~al\mbox{.}(2024)]%
        {luo2024moelora}
\bibfield{author}{\bibinfo{person}{Tongxu Luo}, \bibinfo{person}{Jiahe Lei}, \bibinfo{person}{Fangyu Lei}, \bibinfo{person}{Weihao Liu}, \bibinfo{person}{Shizhu He}, \bibinfo{person}{Jun Zhao}, {and} \bibinfo{person}{Kang Liu}.} \bibinfo{year}{2024}\natexlab{}.
\newblock \showarticletitle{Moelora: Contrastive learning guided mixture of experts on parameter-efficient fine-tuning for large language models}.
\newblock \bibinfo{journal}{\emph{arXiv preprint arXiv:2402.12851}} (\bibinfo{year}{2024}).
\newblock


\bibitem[Nakano et~al\mbox{.}(2021)]%
        {nakano2021webgpt}
\bibfield{author}{\bibinfo{person}{Reiichiro Nakano}, \bibinfo{person}{Jacob Hilton}, \bibinfo{person}{Suchir Balaji}, \bibinfo{person}{Jeff Wu}, \bibinfo{person}{Long Ouyang}, \bibinfo{person}{Christina Kim}, \bibinfo{person}{Christopher Hesse}, \bibinfo{person}{Shantanu Jain}, \bibinfo{person}{Vineet Kosaraju}, \bibinfo{person}{William Saunders}, \bibinfo{person}{Xu Jiang}, \bibinfo{person}{Karl Cobbe}, \bibinfo{person}{Tyna Eloundou}, \bibinfo{person}{Gretchen Krueger}, \bibinfo{person}{Kevin Button}, \bibinfo{person}{Matthew Knight}, \bibinfo{person}{Benjamin Chess}, {and} \bibinfo{person}{John Schulman}.} \bibinfo{year}{2021}\natexlab{}.
\newblock \showarticletitle{WebGPT: Browser-assisted question-answering with human feedback}. In \bibinfo{booktitle}{\emph{arXiv}}.
\newblock


\bibitem[Navon et~al\mbox{.}(2022)]%
        {navon2022multi}
\bibfield{author}{\bibinfo{person}{Aviv Navon}, \bibinfo{person}{Aviv Shamsian}, \bibinfo{person}{Idan Achituve}, \bibinfo{person}{Haggai Maron}, \bibinfo{person}{Kenji Kawaguchi}, \bibinfo{person}{Gal Chechik}, {and} \bibinfo{person}{Ethan Fetaya}.} \bibinfo{year}{2022}\natexlab{}.
\newblock \showarticletitle{Multi-task learning as a bargaining game}.
\newblock \bibinfo{journal}{\emph{arXiv preprint arXiv:2202.01017}} (\bibinfo{year}{2022}).
\newblock


\bibitem[Pan et~al\mbox{.}(2024)]%
        {pan2024lisa}
\bibfield{author}{\bibinfo{person}{Rui Pan}, \bibinfo{person}{Xiang Liu}, \bibinfo{person}{Shizhe Diao}, \bibinfo{person}{Renjie Pi}, \bibinfo{person}{Jipeng Zhang}, \bibinfo{person}{Chi Han}, {and} \bibinfo{person}{Tong Zhang}.} \bibinfo{year}{2024}\natexlab{}.
\newblock \showarticletitle{LISA: Layerwise Importance Sampling for Memory-Efficient Large Language Model Fine-Tuning}.
\newblock \bibinfo{journal}{\emph{arXiv preprint arXiv:2403.17919}} (\bibinfo{year}{2024}).
\newblock


\bibitem[Papineni et~al\mbox{.}(2002)]%
        {papineni2002bleu}
\bibfield{author}{\bibinfo{person}{Kishore Papineni}, \bibinfo{person}{Salim Roukos}, \bibinfo{person}{Todd Ward}, {and} \bibinfo{person}{Wei-Jing Zhu}.} \bibinfo{year}{2002}\natexlab{}.
\newblock \showarticletitle{Bleu: a method for automatic evaluation of machine translation}. In \bibinfo{booktitle}{\emph{Proceedings of the 40th annual meeting of the Association for Computational Linguistics}}. \bibinfo{pages}{311--318}.
\newblock


\bibitem[Pfeiffer et~al\mbox{.}(2020a)]%
        {pfeiffer2020adapterfusion}
\bibfield{author}{\bibinfo{person}{Jonas Pfeiffer}, \bibinfo{person}{Aishwarya Kamath}, \bibinfo{person}{Andreas R{\"u}ckl{\'e}}, \bibinfo{person}{Kyunghyun Cho}, {and} \bibinfo{person}{Iryna Gurevych}.} \bibinfo{year}{2020}\natexlab{a}.
\newblock \showarticletitle{Adapterfusion: Non-destructive task composition for transfer learning}.
\newblock \bibinfo{journal}{\emph{arXiv preprint arXiv:2005.00247}} (\bibinfo{year}{2020}).
\newblock


\bibitem[Pfeiffer et~al\mbox{.}(2020b)]%
        {pfeiffer2020mad}
\bibfield{author}{\bibinfo{person}{Jonas Pfeiffer}, \bibinfo{person}{Ivan Vuli{\'c}}, \bibinfo{person}{Iryna Gurevych}, {and} \bibinfo{person}{Sebastian Ruder}.} \bibinfo{year}{2020}\natexlab{b}.
\newblock \showarticletitle{Mad-x: An adapter-based framework for multi-task cross-lingual transfer}.
\newblock \bibinfo{journal}{\emph{arXiv preprint arXiv:2005.00052}} (\bibinfo{year}{2020}).
\newblock


\bibitem[Raffel et~al\mbox{.}(2020)]%
        {raffel2020exploring}
\bibfield{author}{\bibinfo{person}{Colin Raffel}, \bibinfo{person}{Noam Shazeer}, \bibinfo{person}{Adam Roberts}, \bibinfo{person}{Katherine Lee}, \bibinfo{person}{Sharan Narang}, \bibinfo{person}{Michael Matena}, \bibinfo{person}{Yanqi Zhou}, \bibinfo{person}{Wei Li}, {and} \bibinfo{person}{Peter~J Liu}.} \bibinfo{year}{2020}\natexlab{}.
\newblock \showarticletitle{Exploring the limits of transfer learning with a unified text-to-text transformer}.
\newblock \bibinfo{journal}{\emph{Journal of machine learning research}} \bibinfo{volume}{21}, \bibinfo{number}{140} (\bibinfo{year}{2020}), \bibinfo{pages}{1--67}.
\newblock


\bibitem[Rajbhandari et~al\mbox{.}(2022)]%
        {rajbhandari2022deepspeed}
\bibfield{author}{\bibinfo{person}{Samyam Rajbhandari}, \bibinfo{person}{Conglong Li}, \bibinfo{person}{Zhewei Yao}, \bibinfo{person}{Minjia Zhang}, \bibinfo{person}{Reza~Yazdani Aminabadi}, \bibinfo{person}{Ammar~Ahmad Awan}, \bibinfo{person}{Jeff Rasley}, {and} \bibinfo{person}{Yuxiong He}.} \bibinfo{year}{2022}\natexlab{}.
\newblock \showarticletitle{Deepspeed-moe: Advancing mixture-of-experts inference and training to power next-generation ai scale}. In \bibinfo{booktitle}{\emph{International conference on machine learning}}. PMLR, \bibinfo{pages}{18332--18346}.
\newblock


\bibitem[Sun et~al\mbox{.}(2025)]%
        {sun2025stronger}
\bibfield{author}{\bibinfo{person}{Mengyang Sun}, \bibinfo{person}{Yihao Wang}, \bibinfo{person}{Tao Feng}, \bibinfo{person}{Dan Zhang}, \bibinfo{person}{Yifan Zhu}, {and} \bibinfo{person}{Jie Tang}.} \bibinfo{year}{2025}\natexlab{}.
\newblock \showarticletitle{A Stronger Mixture of Low-Rank Experts for Fine-Tuning Foundation Models}.
\newblock \bibinfo{journal}{\emph{arXiv preprint arXiv:2502.15828}} (\bibinfo{year}{2025}).
\newblock


\bibitem[Team et~al\mbox{.}(2023)]%
        {team2023gemini}
\bibfield{author}{\bibinfo{person}{Gemini Team}, \bibinfo{person}{Rohan Anil}, \bibinfo{person}{Sebastian Borgeaud}, \bibinfo{person}{Jean-Baptiste Alayrac}, \bibinfo{person}{Jiahui Yu}, \bibinfo{person}{Radu Soricut}, \bibinfo{person}{Johan Schalkwyk}, \bibinfo{person}{Andrew~M Dai}, \bibinfo{person}{Anja Hauth}, \bibinfo{person}{Katie Millican}, {et~al\mbox{.}}} \bibinfo{year}{2023}\natexlab{}.
\newblock \showarticletitle{Gemini: a family of highly capable multimodal models}.
\newblock \bibinfo{journal}{\emph{arXiv preprint arXiv:2312.11805}} (\bibinfo{year}{2023}).
\newblock


\bibitem[Vaswani et~al\mbox{.}(2017)]%
        {vaswani2017attention}
\bibfield{author}{\bibinfo{person}{Ashish Vaswani}, \bibinfo{person}{Noam Shazeer}, \bibinfo{person}{Niki Parmar}, \bibinfo{person}{Jakob Uszkoreit}, \bibinfo{person}{Llion Jones}, \bibinfo{person}{Aidan~N Gomez}, \bibinfo{person}{{\L}ukasz Kaiser}, {and} \bibinfo{person}{Illia Polosukhin}.} \bibinfo{year}{2017}\natexlab{}.
\newblock \showarticletitle{Attention is all you need}.
\newblock \bibinfo{journal}{\emph{Advances in neural information processing systems}}  \bibinfo{volume}{30} (\bibinfo{year}{2017}).
\newblock


\bibitem[Wang et~al\mbox{.}(2023d)]%
        {wang2023wavelet}
\bibfield{author}{\bibinfo{person}{Jingyuan Wang}, \bibinfo{person}{Chen Yang}, \bibinfo{person}{Xiaohan Jiang}, {and} \bibinfo{person}{Junjie Wu}.} \bibinfo{year}{2023}\natexlab{d}.
\newblock \showarticletitle{WHEN: A Wavelet-DTW hybrid attention network for heterogeneous time series analysis}. In \bibinfo{booktitle}{\emph{Proceedings of the 29th ACM SIGKDD conference on knowledge discovery and data mining}}. \bibinfo{pages}{2361--2373}.
\newblock


\bibitem[Wang et~al\mbox{.}(2025a)]%
        {wang2025put}
\bibfield{author}{\bibinfo{person}{Maolin Wang}, \bibinfo{person}{Jun Chu}, \bibinfo{person}{Sicong Xie}, \bibinfo{person}{Xiaoling Zang}, \bibinfo{person}{Yao Zhao}, \bibinfo{person}{Wenliang Zhong}, {and} \bibinfo{person}{Xiangyu Zhao}.} \bibinfo{year}{2025}\natexlab{a}.
\newblock \showarticletitle{Put Teacher in Student's Shoes: Cross-Distillation for Ultra-compact Model Compression Framework}. In \bibinfo{booktitle}{\emph{Proceedings of the 31st ACM SIGKDD Conference on Knowledge Discovery and Data Mining V. 2}}. \bibinfo{pages}{4975--4985}.
\newblock


\bibitem[Wang et~al\mbox{.}(2025b)]%
        {wang2025metalora}
\bibfield{author}{\bibinfo{person}{Maolin Wang}, \bibinfo{person}{Xiangyu Zhao}, \bibinfo{person}{Ruocheng Guo}, {and} \bibinfo{person}{Junhui Wang}.} \bibinfo{year}{2025}\natexlab{b}.
\newblock \showarticletitle{MetaLoRA: Tensor-Enhanced Adaptive Low-Rank Fine-Tuning}. In \bibinfo{booktitle}{\emph{2025 IEEE 41st International Conference on Data Engineering (ICDE)}}. IEEE, \bibinfo{pages}{4680--4684}.
\newblock


\bibitem[Wang et~al\mbox{.}(2023e)]%
        {wang2023large}
\bibfield{author}{\bibinfo{person}{Maolin Wang}, \bibinfo{person}{Yao Zhao}, \bibinfo{person}{Jiajia Liu}, \bibinfo{person}{Jingdong Chen}, \bibinfo{person}{Chenyi Zhuang}, \bibinfo{person}{Jinjie Gu}, \bibinfo{person}{Ruocheng Guo}, {and} \bibinfo{person}{Xiangyu Zhao}.} \bibinfo{year}{2023}\natexlab{e}.
\newblock \showarticletitle{Large multimodal model compression via efficient pruning and distillation at AntGroup}.
\newblock \bibinfo{journal}{\emph{arXiv preprint arXiv:2312.05795}} (\bibinfo{year}{2023}).
\newblock


\bibitem[Wang et~al\mbox{.}(2023c)]%
        {wang2023rethinking}
\bibfield{author}{\bibinfo{person}{Xiaolei Wang}, \bibinfo{person}{Xinyu Tang}, \bibinfo{person}{Wayne~Xin Zhao}, \bibinfo{person}{Jingyuan Wang}, {and} \bibinfo{person}{Ji-Rong Wen}.} \bibinfo{year}{2023}\natexlab{c}.
\newblock \showarticletitle{Rethinking the evaluation for conversational recommendation in the era of large language models}.
\newblock \bibinfo{journal}{\emph{arXiv preprint arXiv:2305.13112}} (\bibinfo{year}{2023}).
\newblock


\bibitem[Wang et~al\mbox{.}(2023a)]%
        {wang2023multi}
\bibfield{author}{\bibinfo{person}{Yuhao Wang}, \bibinfo{person}{Ha~Tsz Lam}, \bibinfo{person}{Yi Wong}, \bibinfo{person}{Ziru Liu}, \bibinfo{person}{Xiangyu Zhao}, \bibinfo{person}{Yichao Wang}, \bibinfo{person}{Bo Chen}, \bibinfo{person}{Huifeng Guo}, {and} \bibinfo{person}{Ruiming Tang}.} \bibinfo{year}{2023}\natexlab{a}.
\newblock \showarticletitle{Multi-task deep recommender systems: A survey}.
\newblock \bibinfo{journal}{\emph{arXiv preprint arXiv:2302.03525}} (\bibinfo{year}{2023}).
\newblock


\bibitem[Wang et~al\mbox{.}(2023b)]%
        {wang2023multilora}
\bibfield{author}{\bibinfo{person}{Yiming Wang}, \bibinfo{person}{Yu Lin}, \bibinfo{person}{Xiaodong Zeng}, {and} \bibinfo{person}{Guannan Zhang}.} \bibinfo{year}{2023}\natexlab{b}.
\newblock \showarticletitle{Multilora: Democratizing lora for better multi-task learning}.
\newblock \bibinfo{journal}{\emph{arXiv preprint arXiv:2311.11501}} (\bibinfo{year}{2023}).
\newblock


\bibitem[Wang et~al\mbox{.}(2024)]%
        {wang2024llm4msr}
\bibfield{author}{\bibinfo{person}{Yuhao Wang}, \bibinfo{person}{Yichao Wang}, \bibinfo{person}{Zichuan Fu}, \bibinfo{person}{Xiangyang Li}, \bibinfo{person}{Wanyu Wang}, \bibinfo{person}{Yuyang Ye}, \bibinfo{person}{Xiangyu Zhao}, \bibinfo{person}{Huifeng Guo}, {and} \bibinfo{person}{Ruiming Tang}.} \bibinfo{year}{2024}\natexlab{}.
\newblock \showarticletitle{Llm4msr: An llm-enhanced paradigm for multi-scenario recommendation}. In \bibinfo{booktitle}{\emph{Proceedings of the 33rd ACM International Conference on Information and Knowledge Management}}. \bibinfo{pages}{2472--2481}.
\newblock


\bibitem[Wei et~al\mbox{.}(2021)]%
        {wei2021finetuned}
\bibfield{author}{\bibinfo{person}{Jason Wei}, \bibinfo{person}{Maarten Bosma}, \bibinfo{person}{Vincent~Y Zhao}, \bibinfo{person}{Kelvin Guu}, \bibinfo{person}{Adams~Wei Yu}, \bibinfo{person}{Brian Lester}, \bibinfo{person}{Nan Du}, \bibinfo{person}{Andrew~M Dai}, {and} \bibinfo{person}{Quoc~V Le}.} \bibinfo{year}{2021}\natexlab{}.
\newblock \showarticletitle{Finetuned language models are zero-shot learners}.
\newblock \bibinfo{journal}{\emph{arXiv preprint arXiv:2109.01652}} (\bibinfo{year}{2021}).
\newblock


\bibitem[Yu et~al\mbox{.}(2020)]%
        {yu2020gradient}
\bibfield{author}{\bibinfo{person}{Tianhe Yu}, \bibinfo{person}{Saurabh Kumar}, \bibinfo{person}{Abhishek Gupta}, \bibinfo{person}{Sergey Levine}, \bibinfo{person}{Karol Hausman}, {and} \bibinfo{person}{Chelsea Finn}.} \bibinfo{year}{2020}\natexlab{}.
\newblock \showarticletitle{Gradient surgery for multi-task learning}.
\newblock \bibinfo{journal}{\emph{Advances in Neural Information Processing Systems}}  \bibinfo{volume}{33} (\bibinfo{year}{2020}), \bibinfo{pages}{5824--5836}.
\newblock


\bibitem[Yu et~al\mbox{.}(2025)]%
        {yu2025bigcity}
\bibfield{author}{\bibinfo{person}{Xie Yu}, \bibinfo{person}{Jingyuan Wang}, \bibinfo{person}{Yifan Yang}, \bibinfo{person}{Qian Huang}, {and} \bibinfo{person}{Ke Qu}.} \bibinfo{year}{2025}\natexlab{}.
\newblock \showarticletitle{BIGCity: A universal spatiotemporal model for unified trajectory and traffic state data analysis}. In \bibinfo{booktitle}{\emph{2025 IEEE 41st International Conference on Data Engineering (ICDE)}}. IEEE, \bibinfo{pages}{4455--4469}.
\newblock


\bibitem[Zaken et~al\mbox{.}(2021)]%
        {zaken2021bitfit}
\bibfield{author}{\bibinfo{person}{Elad~Ben Zaken}, \bibinfo{person}{Shauli Ravfogel}, {and} \bibinfo{person}{Yoav Goldberg}.} \bibinfo{year}{2021}\natexlab{}.
\newblock \showarticletitle{Bitfit: Simple parameter-efficient fine-tuning for transformer-based masked language-models}.
\newblock \bibinfo{journal}{\emph{arXiv preprint arXiv:2106.10199}} (\bibinfo{year}{2021}).
\newblock


\bibitem[Zhang et~al\mbox{.}(2023a)]%
        {zhang2023adalora}
\bibfield{author}{\bibinfo{person}{Qingru Zhang}, \bibinfo{person}{Minshuo Chen}, \bibinfo{person}{Alexander Bukharin}, \bibinfo{person}{Nikos Karampatziakis}, \bibinfo{person}{Pengcheng He}, \bibinfo{person}{Yu Cheng}, \bibinfo{person}{Weizhu Chen}, {and} \bibinfo{person}{Tuo Zhao}.} \bibinfo{year}{2023}\natexlab{a}.
\newblock \showarticletitle{AdaLoRA: Adaptive budget allocation for parameter-efficient fine-tuning}.
\newblock \bibinfo{journal}{\emph{arXiv preprint arXiv:2303.10512}} (\bibinfo{year}{2023}).
\newblock


\bibitem[Zhang et~al\mbox{.}(2024)]%
        {zhang2024veccity}
\bibfield{author}{\bibinfo{person}{Wentao Zhang}, \bibinfo{person}{Jingyuan Wang}, \bibinfo{person}{Yifan Yang}, {et~al\mbox{.}}} \bibinfo{year}{2024}\natexlab{}.
\newblock \showarticletitle{VecCity: A taxonomy-guided library for map entity representation learning}.
\newblock \bibinfo{journal}{\emph{arXiv preprint arXiv:2411.00874}} (\bibinfo{year}{2024}).
\newblock


\bibitem[Zhang et~al\mbox{.}(2023b)]%
        {zhang2023automatic}
\bibfield{author}{\bibinfo{person}{Zhuosheng Zhang}, \bibinfo{person}{Aston Zhang}, \bibinfo{person}{Mu Li}, {and} \bibinfo{person}{Alex Smola}.} \bibinfo{year}{2023}\natexlab{b}.
\newblock \showarticletitle{Automatic Chain of Thought Prompting in Large Language Models}. In \bibinfo{booktitle}{\emph{The Eleventh International Conference on Learning Representations (ICLR 2023)}}.
\newblock


\bibitem[Zhao et~al\mbox{.}(2023)]%
        {zhao2023survey}
\bibfield{author}{\bibinfo{person}{Wayne~Xin Zhao}, \bibinfo{person}{Kun Zhou}, \bibinfo{person}{Junyi Li}, \bibinfo{person}{Tianyi Tang}, \bibinfo{person}{Xiaolei Wang}, \bibinfo{person}{Yupeng Hou}, \bibinfo{person}{Yingqian Min}, \bibinfo{person}{Beichen Zhang}, \bibinfo{person}{Junjie Zhang}, \bibinfo{person}{Zican Dong}, {et~al\mbox{.}}} \bibinfo{year}{2023}\natexlab{}.
\newblock \showarticletitle{A survey of large language models}.
\newblock \bibinfo{journal}{\emph{arXiv preprint arXiv:2303.18223}} (\bibinfo{year}{2023}).
\newblock


\end{thebibliography}


\end{document}